\documentclass{applemlr}

\usepackage{amsmath}
\usepackage{enumerate}
\usepackage{algorithm}
\usepackage{algpseudocode}
\usepackage{amsfonts}
\usepackage{amsthm}
\usepackage{cleveref}

\theoremstyle{plain}

\theoremstyle{definition}

\makeatletter
\let\c@proposition\c@theorem 
\let\c@lemma\c@theorem       
\let\c@corollary\c@theorem   
\let\c@definition\c@theorem  
\let\c@assumption\c@theorem  
\makeatother

\usepackage{diagbox}
\usepackage{colortbl}
\usepackage{amssymb}
\usepackage{xspace}
\usepackage{wrapfig}
\usepackage{adjustbox}
\usepackage{tabularx}
\usepackage{booktabs}
\usepackage{mathtools}
\usepackage{tikz}
\usepackage{enumitem}
\usepackage{silence}
\usepackage{dsfont}
\usepackage[table]{xcolor}
\usepackage[dvipsnames]{xcolor}
\usepackage{multirow}
\usepackage{makecell}
\usepackage{xfakebold}

\usepackage{amsmath,amsfonts,bm}

\def\eqref#1{equation~\ref{#1}}

\def\1{\bm{1}}

\DeclareMathAlphabet{\mathsfit}{\encodingdefault}{\sfdefault}{m}{sl}
\SetMathAlphabet{\mathsfit}{bold}{\encodingdefault}{\sfdefault}{bx}{n}

\definecolor{textgray}{HTML}{6E6E73}
\usetikzlibrary{positioning, calc}
\usetikzlibrary{decorations.pathmorphing}

\makeatletter
\patchcmd{\wrong@fontshape}{\@gobbletwo}{}{}{}
\makeatother
\numberwithin{equation}{section}
\makeatletter
\AtBeginDocument{
  \urlstyle{sf}
  
}
\makeatother

\definecolor{light}{RGB}{125, 125, 125}
\definecolor{promptframe}{HTML}{2F4858}
\definecolor{promptback}{HTML}{F7F8FA}
\definecolor{promptrule}{HTML}{D6DBE0}
\crefname{tcb@cnt@pbox}{code}{code}
\Crefname{tcb@cnt@pbox}{Code}{Code}
\crefname{assumption}{assumption}{assumption}
\Crefname{assumption}{Assumption}{Assumptions}
\crefname{proposition}{Proposition}{Propositions}
\Crefname{proposition}{Proposition}{Propositions}
\crefname{corollary}{Corollary}{Corollaries}
\Crefname{corollary}{Corollary}{Corollaries}
\crefname{lemma}{Lemma}{Lemmas}
\Crefname{lemma}{Lemma}{Lemmas}
\crefname{definition}{Definition}{Definitions}
\Crefname{definition}{Definition}{Definitions}

\tcbuselibrary{skins,breakable}
\newtcolorbox[auto counter]{pbox}[2][]{
  breakable,
  enhanced,
  colback=promptback,
  title={Prompt~\thetcbcounter\;\; #2},
  #1,fonttitle=\sffamily\bfseries\small,
  before upper={\raggedright\hyphenpenalty=10000\exhyphenpenalty=10000},
  arc=2.5pt,
  boxrule=0.5pt,
  colframe=promptframe,
  coltitle=white,
  colbacktitle=promptframe,
  toptitle=1.4mm,
  bottomtitle=1.4mm,
  left=3.2mm,right=3.2mm,top=2.6mm,bottom=2.6mm
}

\makeatletter
\newcommand\applefootnote[1]{%
  \begingroup
  \renewcommand\thefootnote{}%
  \renewcommand\@makefntext[1]{\noindent##1}%
  \footnote{#1}%
  \addtocounter{footnote}{-1}%
  \endgroup
}
\makeatother

\definecolor{cverbbg}{gray}{0.90}

\usepackage{url}
\usepackage{float}
\usepackage{placeins}
\usepackage{needspace}
\usepackage{array}
\usepackage{pifont}
\usepackage{pdflscape}
\usepackage{titletoc}
\usepackage{arydshln}
\usepackage[colorinlistoftodos]{todonotes}
\usetikzlibrary{arrows.meta}
\titlespacing*{\paragraph}{0pt}{1.5ex plus .5ex minus .2ex}{0.8em}

\crefname{tcb@cnt@pbox}{Prompt}{Prompts}
\Crefname{tcb@cnt@pbox}{Prompt}{Prompts}

\definecolor{pboxaccent}{HTML}{1F4E9C}
\newcommand{\tvar}[1]{\textcolor{pboxaccent}{\texttt{\{\{\,#1\,\}\}}}}
\newcommand{\btag}[1]{\textcolor{pboxaccent!80!black}{\texttt{#1}}}

\definecolor{takeawaybg}{HTML}{EFF4FB}
\definecolor{takeawayrule}{HTML}{1F4E9C}
\newtcolorbox{takeaway}{
  enhanced,
  colback=takeawaybg,
  colframe=takeawayrule,
  boxrule=0.5pt,
  arc=2pt,
  left=6pt, right=6pt, top=3pt, bottom=3pt,
  before skip=6pt, after skip=6pt,
  fontupper=\small,
}
\newcommand{\takeawaytext}[1]{\begin{takeaway}\textbf{Takeaway.} #1\end{takeaway}}
\newcommand{\pslot}[1]{\textcolor{black!55}{$\langle$\,\textit{#1}\,$\rangle$}}

\newcommand{\truncname}{\texttt{Trunc}}
\newcommand{\gatedname}{\texttt{Gate}}
\newcommand{\magname}{\texttt{Linear-Mag}}
\newcommand{\ratename}{\texttt{Rate}}
\newcommand{\baponame}{\texttt{Dual}}
\newcommand{\gatenegname}{\texttt{Gate-Neg}}
\newcommand{\costname}{\texttt{Linear-Cost}}
\newcommand{\expmagname}{\texttt{Exp-Mag}}
\newcommand{\sigmoidgatename}{\texttt{Sigmoid-Gate}}
\newcommand{\sigmoidname}{\texttt{Sigmoid}}
\newcommand{\basename}{\texttt{Base}}
\newcommand{\cmark}{\ding{51}}
\newcommand{\xmark}{\textcolor{gray}{\ding{55}}}

\newcommand{\grponormalname}{\texttt{GRPO-Std}}

\title{Rewards for Budget-Aware Multi-Turn Agents That Adapt to Arbitrary Test-Time Constraints}

\author[1,2,*]{Tom A. Lamb}
\author[1]{Adam Golinski}
\author[1]{Eleonora Gualdoni}
\author[1]{Sinead Williamson}
\author[1]{Michael Kirchhof}

\affiliation[1]{Apple}
\affiliation[2]{University of Oxford}

\contribution[*]{Work completed during an internship at Apple.}

\abstract{Large language model agents are often given an inference budget and asked to maximise success within it. A multi-turn agent has to allocate this budget across many environment interactions. An external harness can enforce the budget, but it remains unclear whether the agent itself can learn when and how much to spend. We ask whether agents can learn this allocation across dynamically specified budgets and which rewards best teach this behaviour. We adapt existing length penalties to budget constraints spanning multiple resource types, introduce new reward formulations, and compare eleven objectives in a shared training pipeline. We train Qwen3.5 models at 4B and 9B on AppWorld with budgets over tokens, API calls, and environment steps, and on the BabyAI gridworld with turn budgets. We train a separate policy for each environment and resource type, with each policy trained across budget values. Training with rewards that penalise overspending substantially improves within-budget success and natural compliance over untrained models and over training on task success alone, with agents learning to adapt their resource use to stated budgets and task difficulty. These gains extend to unseen budgets and, zero-shot, to a shifted task distribution and a new benchmark. Reward objectives trade off success within budget and natural compliance. A sigmoid penalty for overspending most often achieves the highest within-budget success at an individual budget, while a fixed negative reward for overspending generally yields the strongest natural compliance. Combining penalties for overspend frequency and magnitude gives the smallest average gap to the best method at each budget. Overall, we show that a single agent can learn to allocate resources across turns and adapt its behaviour to user-specified budgets using budget feedback without external budget-allocation controllers.}

\metadata[Correspondence]{\sffamily Tom A. Lamb: \url{thomas.lamb@eng.ox.ac.uk}}

\date{\sffamily\today}

\begin{document}

\maketitle

\section{Introduction}
\label{sec:intro}

Agents can solve harder tasks by spending more at test time, thinking for longer and taking more actions in an environment \citep{snell2024scaling, guo2025deepseek, yu2025dapo}. 
Current frontier agents give users some control over the effort in terms of generated tokens, tool calls, or environment actions \citep{liu2025budget}.
To respond to different budgets, an agent needs to adapt its strategy to maximise success under the allowance it receives.

Budget-aware training has largely focused on controlling the cost of a single model generation \citep{aggarwal2025l1, liu2026laconic, wen2025budgetthinker}.
A multi-turn agent instead shares one budget across many turns, and at each turn the policy has to decide when and how much to spend given what remains \citep{liu2025budget}. This allocation is usually left to external machinery at inference time, including planners, harnesses, or routers, and is not learned by the policy itself \citep{zhang2026budget, liu2025budget}.
That machinery adds complexity at inference, and it remains unclear whether the task-performing policy can learn to allocate these resources itself.
Different constraints may also call for different behaviour. This leads to our first question. \textit{Can we train a multi-turn agent to maximise task success under dynamically specified budgets, across different inference resources?}

We study this through reward shaping. In single-generation tasks, length penalties are often subtracted from rewards when an agent exceeds a token limit. Overlong trajectories are still collected, and they can show how to solve a task, which helps most when successes within budget are rare. Some frontier models use linear penalties \citep{yu2025dapo,deepseek2026v41flash}, and others use step functions \citep{team2026kimi}. Learned budget penalties draw on constrained optimisation \citep{altman1999constrained, stooke2020responsive}, while cost penalties have also been used to train budget-aware routers \citep{zhang2026budget}. It is not clear in advance which objectives best train a single multi-turn policy across dynamically specified budgets and different resource types. This leads to our second question. \textit{Which training objectives best teach multi-turn agents to maximise task success while complying with dynamically specified budgets?}

\begin{figure}[t]
\centering
\vspace{-0pt}
\sbox0{\includegraphics[width=0.63\textwidth]{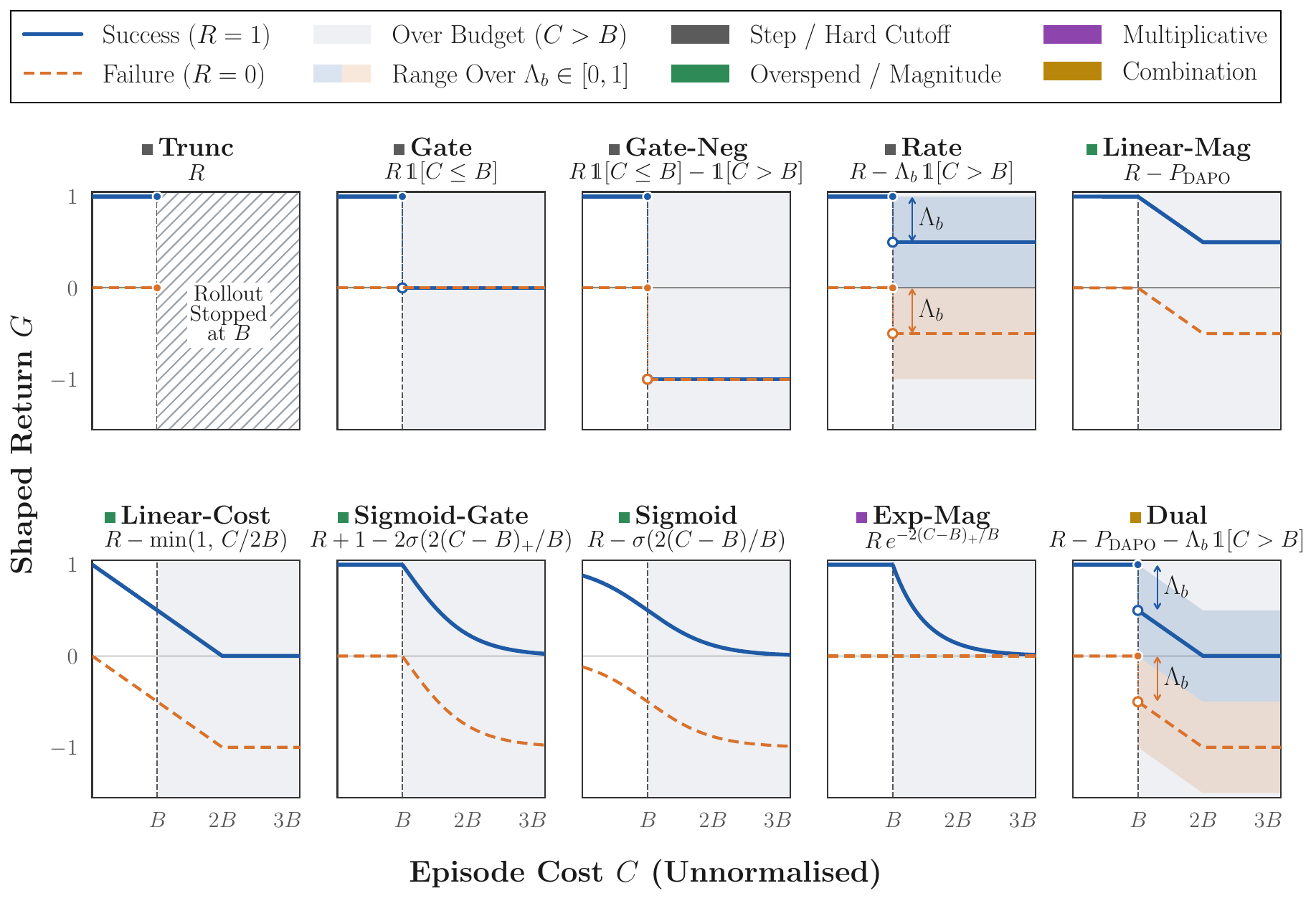}}%
\usebox0\hfill
\resizebox{!}{\ht0}{%
\scriptsize
\setlength{\tabcolsep}{3.5pt}
\renewcommand{\arraystretch}{1.3}
\begin{tabular}[b]{@{}lcccccc@{}}
\toprule
Reward & \rotatebox{90}{Budget Cutoff} &
\rotatebox{90}{Budget Jump} & \rotatebox{90}{In-Budget Cost} &
\rotatebox{90}{Overspend Size} &
\rotatebox{90}{Failure Return $<0$} & \rotatebox{90}{Learned Price} \\
\midrule
\truncname{} & \cmark & \xmark & \xmark & \xmark & \xmark & \xmark \\
\gatedname{} & \xmark & \cmark & \xmark & \xmark & \xmark & \xmark \\
\gatenegname{} & \xmark & \cmark & \xmark & \xmark & \cmark & \xmark \\
\ratename{} & \xmark & \cmark & \xmark & \xmark & \cmark & \cmark \\
\magname{} & \xmark & \xmark & \xmark & \cmark & \cmark & \xmark \\
\costname{} & \xmark & \xmark & \cmark & \cmark & \cmark & \xmark \\
\sigmoidgatename{} & \xmark & \xmark & \xmark & \cmark & \cmark & \xmark \\
\sigmoidname{} & \xmark & \xmark & \cmark & \cmark & \cmark & \xmark \\
\expmagname{} & \xmark & \xmark & \xmark & \cmark & \xmark & \xmark \\
\baponame{} & \xmark & \cmark & \xmark & \cmark & \cmark & \cmark \\
\bottomrule
\end{tabular}}
\caption{\textbf{Cost-conditioned rewards.}
[Left] Return $G$ against episode cost $C$ for successful ($R{=}1$) and failed ($R{=}0$) trajectories across different cost-conditional rewards. Shading marks $C>B$. Bands illustrate $\Lambda_b\in[0,1]$ for \ratename{} and \baponame{}.
[Right] Reward properties: \emph{Budget Cutoff} stops rollouts before exceeding $B$, \emph{Budget Jump} denotes a return discontinuity at $B$, \emph{In-Budget Cost} penalises resource use before $B$, \emph{Overspend Size} makes the return depend on the extent of overspending, \emph{Failure Return $<0$} means failed trajectories can receive negative return, and \emph{Learned Price} denotes a separate learned violation price for each training budget.}
\label{fig:reward-functions}
\end{figure}

To address these two questions, we introduce a framework for training a single policy across dynamically specified episode-level budget values, with a separate policy for each environment and resource type. The policy is told its total and remaining budget in natural language at every turn, so it can learn to allocate resources as the episode goes on.
Within this framework, we adapt existing penalised objectives that punish excessive resource use or budget violations to the multi-turn, multi-budget setting and introduce new reward formulations. We then compare these objectives directly to see which ones best teach budget-aware behaviour. We train and evaluate on AppWorld \citep{trivedi2024appworld}, with budgets on generated tokens, environment interactions, or tool calls. We test generalisation to budgets not seen during training, distribution shift on AppWorld \texttt{test-challenge}, and zero-shot transfer to ASTRA \citep{xiu2026astra}. To broaden the evaluation to a different environment and task format, we also train and evaluate on BabyAI \citep{chevalier-boisvert2019babyai}, a text-based gridworld.

Our contributions are as follows:
\begin{itemize}[leftmargin=*, topsep=0pt, itemsep=7pt, partopsep=0pt, parsep=0pt]
\item We introduce a framework for training and evaluating multi-turn agents across dynamically specified budgets and resource types, with metrics for task success within budget and natural budget compliance.

\item We adapt existing cost-conditioned objectives to the multi-turn, multi-budget setting and introduce new reward formulations, covering hard success gating, continuous cost shaping, and fixed or learned overspend penalties.

\item We show that the strongest budget-shaped objectives improve within-budget success and natural compliance over untrained models and budget-independent task-reward training across token, tool-call, and environment-step budgets. On AppWorld, learned resource allocation responds to the stated budget and task difficulty and generalises to unseen budget values. Under step budgets, improvements persist on \texttt{test-challenge} and under zero-shot transfer to ASTRA. We also evaluate budget-aware behaviour in BabyAI.

\item We identify distinct trade-offs between budget utility, natural compliance, and overspend severity, with no single objective dominating across all criteria. \baponame{} has the smallest average utility gap to the best method at each budget, \sigmoidgatename{} most often achieves the highest utility at an individual budget, and \gatenegname{} generally achieves the highest natural compliance.
\end{itemize}

\section{Related Work}
\label{sec:related}

\paragraph{Token budgets.}
On-policy budget-aware training without an external harness has mostly controlled the length of a single model generation, rather than how resources are allocated across a multi-turn episode. Existing methods optimise adherence to token budgets \citep{aggarwal2025l1, liu2026laconic, han-etal-2025-token} or learn levels of reasoning effort \citep{arora2025training}, largely in mathematical reasoning. In these settings, the model controls resource use within one generation, and tokens are the resource of interest. A multi-turn agent faces a different allocation problem: it must distribute a shared budget across successive generations and environment interactions, deciding how much to spend at each turn to maximise task success. The constrained resource may also be tool calls or environment steps, rather than just tokens. 

\citet{wen2025budgetthinker} trains across a curriculum of token budgets and uses control tokens inserted during decoding to signal the remaining budget. DAPO \citep{yu2025dapo} introduces an overspend penalty for a fixed generation-length budget. Kimi K3 \citep{team2026kimi} handles several effort levels in agentic domains by training a separate expert for each with step-function reward shaping and then distilling them. We study how to train one policy across dynamically specified episode-level budget values, with separate training for each environment and resource type.
\vspace{1em}
\paragraph{Budget feedback and allocation.} Budget-aware behaviour can also be added without training it into the task-performing policy. \citet{liu2025budget} show that raising a search agent's tool-call budget does not by itself improve results, while reporting the total and remaining budget to the agent can improve scaling for large models when combined with planning and verification strategies.
In our evaluation, the untrained 4B and 9B models struggle to succeed within budget despite receiving total and remaining budget feedback (\Cref{fig:appworld-main}).
Other approaches predict how much compute a task requires \citep{damani2025learninghard} or route between cheaper and more expensive models under a budget \citep{zhang2026budget}.
\citet{li2026spend} search a value tree at inference time on multi-hop question answering and scale node values by the remaining resource ratio, so the search moves from exploration to exploitation as the budget runs down. These methods all handle allocation in planning, harnesses, or routers at inference time. We instead train the task-performing policy to condition its actions on the total and remaining budget, learning to allocate resources effectively across turns.
\vspace{1em}
\paragraph{Constrained decision-making and budgeted MDPs.} Constrained Markov decision processes \citep[cMDPs;][]{altman1999constrained, achiam2017cpo} typically bound expected cumulative cost and can be optimised using Lagrangian relaxation \citep{stooke2020responsive}. If the constraints are pooled across the training distribution, a policy can break tight budgets and make up for it on generous ones. Budgeted MDPs \citep{carrara2019budgeted} extend this to families of budgets by conditioning on the remaining budget allowance, and chance-constrained formulations \citep{geibel2005risk, haskell2015convex, chow2017risk} bound violation probabilities directly. We carry these formulations over to multi-turn LLM agent training with a separate rate constraint for each budget, so the constrained objective treats every budget separately and does not pool compliance across budgets.

\Needspace{12\baselineskip}
\section{Budget-Aware Policy Optimisation}
\label{sec:method}

\subsection{Preliminaries}
\label{sec:prelim}

\begin{wrapfigure}{r}{0.4\linewidth}
  \vspace{-32pt}
  \centering
  \includegraphics[width=\linewidth]{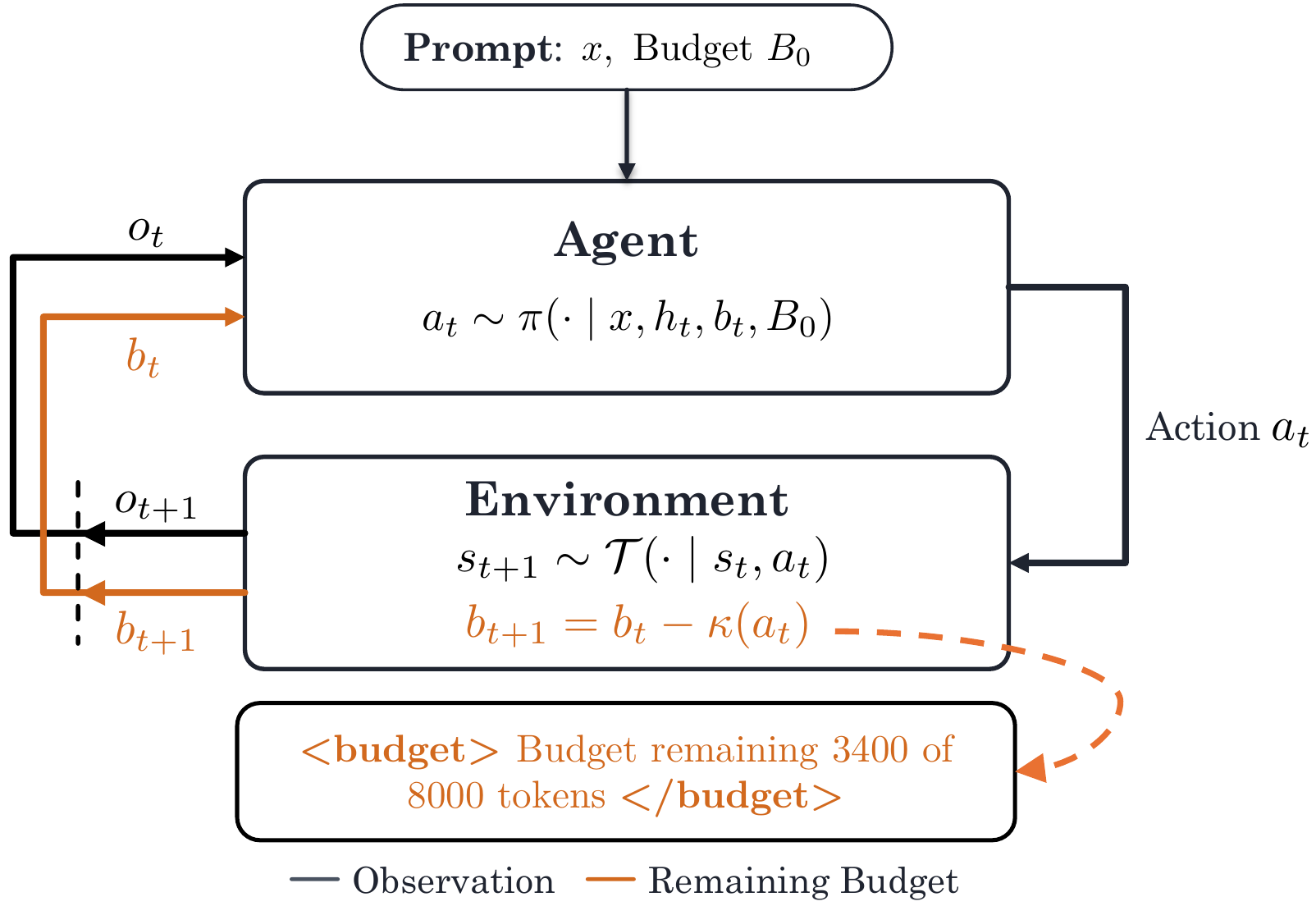}
  \captionsetup{skip=2pt}
\caption{\textbf{Budget-augmented POMDP.} The policy receives the task $x$, total budget $B_0$, and remaining budget $b_t$ alongside the environment observation $o_t$. Each action's cost is deducted from $b_t$, and the updated budget $b_{t+1}$ is appended in natural language to the next observation (\Cref{sec:bapomdp}).}
  \label{fig:budget-pomdp}
  \vspace{-12pt}
\end{wrapfigure}
We write $\mathcal{P}(\mathcal{Y})$ for the set of probability measures over a set $\mathcal{Y}$, and model a task as a POMDP $\mathcal{M} = \langle \mathcal{S}, \mathcal{A}, \mathcal{T}, \Omega, \mathcal{O}, R, \gamma, \mu_0 \rangle$. Here $\mathcal{S}$, $\mathcal{A}$ and $\Omega$ are the state, action and observation spaces. The transition and observation kernels are $\mathcal{T} : \mathcal{S} \times \mathcal{A} \to \mathcal{P}(\mathcal{S})$ and $\mathcal{O} : \mathcal{S} \times \mathcal{A} \to \mathcal{P}(\Omega)$, with $R$ the reward, which \Cref{sec:bapomdp} specialises to a binary terminal reward, $\gamma \in (0, 1]$ the discount factor and $\mu_0 \in \mathcal{P}(\mathcal{S})$ the initial state distribution. Tasks $x \sim p_{\mathcal{X}} \in \mathcal{P}(\mathcal{X})$ are drawn from a distribution over the task set $\mathcal{X}$. At turn $t$, the agent receives the observation history $h_t := (o_1, a_1, \dots, a_{t-1}, o_t)$, comprising previous actions and the resulting environment observations. Writing $\mathcal{H}$ for the set of all such histories, the agent acts according to a policy $\pi : \mathcal{X} \times \mathcal{H} \to \mathcal{P}(\mathcal{A})$.

We optimise policies with Dr.\ GRPO \citep{liu2025understanding}, a variant of group relative policy optimisation (GRPO, \citealp{shao2024deepseekmath}) that baselines each rollout against its group mean and removes response-length and reward-standard-deviation normalisation (\Cref{app:grpo}). Removing these normalisers addresses biases that can encourage failed rollouts to grow longer. The practical training pipeline is described in \Cref{app:train-grpo}.

\subsection{Budget-Augmented POMDP}
\label{sec:bapomdp}

\paragraph{Budget augmentation.}
We augment the task POMDP $\mathcal{M}$ of \Cref{sec:prelim} by pairing the environment state $s_t$ with a scalar remaining budget $b_t$, following the state augmentation used in budgeted MDPs \citep{carrara2019budgeted}. This gives the \emph{budget-augmented} POMDP (\Cref{fig:budget-pomdp}) $\tilde{\mathcal{M}} = \langle \tilde{\mathcal{S}}, \mathcal{A}, \tilde{\mathcal{T}}, \tilde{\Omega}, \tilde{\mathcal{O}}, R, \gamma, \tilde{\mu}_0 \rangle$, with states $\tilde{s}_t = (s_t, b_t)$ and state space $\tilde{\mathcal{S}} := \mathcal{S} \times \mathbb{R}$. Resource use is measured by a cost map $\kappa : \mathcal{S} \times \mathcal{A} \times \mathcal{S} \to \mathbb{R}_{\geq 0}$. Each action is the whole string the policy emits at a turn, including thinking tokens. Its step cost is one, its token cost is its length in tokens, and its API-call cost counts the calls actually executed. These raw costs are independent of the remaining budget. We abbreviate the realised transition cost as $\kappa(a_t) = \kappa(s_t,a_t,s_{t+1}) \in \mathbb{R}_{\geq 0}$. The remaining budget informs the policy's action selection. After each turn, we subtract the realised cost from the remaining budget:
\begin{equation}
b_{t+1} = b_t - \kappa(a_t).
\label{eq:budget-update}
\end{equation}
The augmented transition kernel combines the unchanged environment dynamics with this budget update.
\begin{equation}
\tilde{\mathcal{T}}\big(s_{t+1}, b_{t+1} \mid s_t, b_t, a_t\big)
=
\mathcal{T}\big(s_{t+1} \mid s_t, a_t\big)\,
\mathbf{1}\big[\, b_{t+1} = b_t - \kappa(a_t) \,\big].
\label{eq:transition}
\end{equation}

\paragraph{Budget-conditioned policy.}
The initial budget $B_0 \in \mathbb{R}_{>0}$ and the remaining budget $b_t$ are written into the text of the observations the agent sees (\Cref{app:data-prompts}). At the next turn, the augmented observation is $\tilde{o}_{t+1} = (o_{t+1}, b_{t+1}) \in \tilde{\Omega} := \Omega \times \mathbb{R}$, where $b_{t+1}$ is given by \Cref{eq:budget-update}. The observation kernel preserves the task observation and reveals the updated budget exactly.
\begin{equation}
\tilde{\mathcal{O}}\big(o_{t+1}, b \mid s_{t+1}, b_{t+1}, a_t\big)
=
\mathcal{O}\big(o_{t+1} \mid s_{t+1}, a_t\big)\,
\mathbf{1}\big[\, b = b_{t+1} \,\big].
\label{eq:observation}
\end{equation}
An augmented history splits into the history $h_t$ of $\mathcal{M}$ and the observed budgets,
so the \emph{budget-conditioned} policy
\begin{equation}
\pi(\cdot \mid x, h_t, b_t, B_0) \in \mathcal{P}(\mathcal{A}),
\qquad
\pi : \mathcal{X} \times \mathcal{H} \times \mathbb{R} \times \mathbb{R}_{>0} \to \mathcal{P}(\mathcal{A}),
\label{eq:policy}
\end{equation}
can weigh both the allowance left and the fraction $b_t / B_0$ still available against what is needed to continue toward successfully solving a task from a given state. We write $\Pi$ for the set of all such policies. Throughout post-training the budget is an explicit input to the policy. Since \Cref{eq:policy} conditions directly on $B_0$ and $b_t$, which are written into the environment observations, a budget outside the training grid needs no change to the policy or to inference, and \Cref{sec:experiments} shows that budget-aware behaviour carries over to such unseen budgets and to new environments zero-shot.

\paragraph{Episode termination.}
Each episode draws a task $x \!\sim\! p_{\mathcal{X}}$ and, independently of it, a budget $B_0 \!\sim\! p_B$ from the budget prior defined below, initialising the episode at $b_1 \!=\! B_0$ with the opening observation $o_1$ the prompt stating the task and its budget (see \Cref{app:data-prompts} for the prompts in full). Most methods allow $b_t\!<\!0$, so the agent can overspend and learn from doing so, possibly with a penalty.
In AppWorld and ASTRA, a special \texttt{complete\_task} action $a_\perp \!\in\! \mathcal{A}$ lets the agent end the episode and has cost $\kappa(a_\perp) \!>\! 0$. An episode can also end through environment termination, as when the agent completes a BabyAI mission, or at a finite horizon $T_{\max} \!\in\! \mathbb{Z}_{>0}$. Writing $\tau_{\mathrm{env}}$ for the first turn at which the environment terminates, the stopping time is $\tau \!:=\! \min\big( \inf\{ t \!\geq\! 1 : a_t \!=\! a_\perp \},\ \tau_{\mathrm{env}},\ T_{\max} \big)$, with unavailable termination conditions assigned $+\infty$.

\paragraph{Reward and budget penalties.}
We score the completed trajectory $\zeta := (o_1, a_1, \dots, o_\tau, a_\tau, o_{\tau+1})$ by a binary task reward $R(\zeta) \in \{0,1\}$ and transform it into a budget-conditioned return $\tilde{R}(\zeta, B_0) \in \mathbb{R}$. The agent is trained to maximise the expected shaped return
\begin{equation}
\max_{\pi \in \Pi}\ \ \mathbb{E}_{\substack{x \sim p_{\mathcal{X}},\ B_0 \sim p_B , \zeta \sim \pi(\cdot \mid x, B_0)}}\!\big[\, \tilde{R}(\zeta, B_0) \,\big]\,.
\label{eq:budget-objective}
\end{equation}

We do not discount, so any pressure to finish early comes from the budget and any overspending shaping or penalties. We fix a finite set of $M \in \mathbb{Z}_{>0}$ \emph{training} budgets $\mathcal{B} = \{ b^{(1)} < \dots < b^{(M)} \} \subset \mathbb{R}_{>0}$ and draw budgets uniformly at random during training. The raw spend $C(\zeta) := \sum_{t \leq \tau} \kappa(a_t) \in \mathbb{R}_{\geq 0}$ is always the budgeted quantity, e.g.\ steps, tool calls, or tokens, consumed over the entire multi-step trajectory. Rewards are shaped as a function of this spend as a fraction of the prompted budget $B_0$, the normalised cost
\begin{equation}
\bar{C}(\zeta, B_0) := \frac{C(\zeta)}{B_0} = \frac{1}{B_0} \sum_{t \leq \tau} \kappa(a_t) \in \mathbb{R}_{\geq0},
\label{eq:normcost}
\end{equation}
with  $\bar{C} \leq 1$  within budget and  $\bar{C}>1$ over budget.

\subsection{Cost-Conditioned Rewards}
\label{sec:shaped}

We compare existing reward objectives and new formulations motivated by heuristics, common practice, or constrained optimisation. We organise them into four groups: (a) hard boundary mechanisms that act at the budget limit, (b) within-budget cost pressure that penalises resource use before the limit, (c) gradual post-budget shaping that penalises the amount of overspend, and (d) learned per-budget violation prices that adapt penalties to the overspend rate at each training budget. The objectives also differ in whether overspending can produce negative returns. \Cref{fig:reward-functions} and \Cref{app:rewards} give their mathematical definitions and further details.

\paragraph{Hard boundary mechanisms.}
At the budget, these methods either stop the rollout or assign a fixed outcome to a violation.
\begin{itemize}[leftmargin=1em, topsep=2pt, itemsep=5pt, partopsep=0pt, parsep=0pt]
\item \textbf{\truncname{}} \citep{liu2025dlerdoinglengthpenalty} is a baseline that does not permit any overspending. Every rollout is terminated at $B_0$, and the reward is the binary task reward obtained within $B_0$.
\item \textbf{\gatedname{}} relaxes this constraint. The policy may overspend, but any rollout that exceeds $B_0$ receives zero reward regardless of task success. This provides a simple objective that rewards only success within budget.
\item \textbf{\gatenegname{}} \citep{team2026kimi} assigns a reward of $-1$ whenever the policy exceeds $B_0$, regardless of task success. Unlike \gatedname{}, it therefore prefers a failure within budget to a failure over budget.
\end{itemize}

\paragraph{Within-budget cost pressure.}
These methods begin penalising cost before the budget is exhausted.
\begin{itemize}[leftmargin=1em, topsep=2pt, itemsep=5pt, partopsep=0pt, parsep=0pt]
\item \textbf{\sigmoidname{}} smooths the hard overspend indicator that \ratename{} and \baponame{} apply as a step into a symmetric sigmoid penalty, which applies pressure to solve successfully within budget from the start of a policy's rollout.
\item \textbf{\costname{}} \citep{deepseek2026v41flash} penalises cost linearly from the first action onwards, so the pressure starts well before the budget is reached.
\end{itemize}

\paragraph{Gradual post-budget shaping.}
These methods preserve the full in-budget return, then change it continuously with the amount of overspend.
\begin{itemize}[leftmargin=1em, topsep=2pt, itemsep=5pt, partopsep=0pt, parsep=0pt]
\item \textbf{\sigmoidgatename{}} subtracts a sigmoid-shaped penalty only after $B_0$. It leaves the in-budget return unchanged and continuously lowers the return of both successes and failures as overspend grows. Unlike \gatedname{}, it can assign negative returns to overspending failures.
\item \textbf{\magname{}} has no step at $B_0$ and instead reduces the reward linearly beyond it. It was popularised by DAPO \citep{yu2025dapo}, and we generalise it to multiple budgets in \Cref{app:dapo}.
\item \textbf{\expmagname{}} decays a successful trajectory's reward exponentially in how far it overspends. Failures always receive a reward of $0$, regardless of spend.
\end{itemize}

\paragraph{Learned per-budget violation prices.}
These methods learn an overspend price separately at each budget.
\begin{itemize}[leftmargin=1em, topsep=2pt, itemsep=5pt, partopsep=0pt, parsep=0pt]
\item \textbf{\ratename{}} tracks the overspend rate at each budget and applies a larger penalty the more often the policy exceeds that budget. We derive it from a Lagrangian perspective in \Cref{app:rewards}. Its price is capped at $1.0$, so an overspending success never scores below a compliant failure.
\item \textbf{\baponame{}} combines a step penalty with a linear one. A learned Lagrangian step penalty prices \emph{how often} rollouts at the given budget overspend, and a linear penalty prices \emph{how much} the current rollout overspends. Each term subtracts at most $0.5$. We study the dual optimisation in \Cref{app:optimisation} and why the two penalties are complementary in \Cref{app:roles}.
\end{itemize}

\FloatBarrier

\section{Experiments}
\label{sec:experiments}
\begin{figure}[t]
    \centering
    \includegraphics[width=1.0\linewidth]{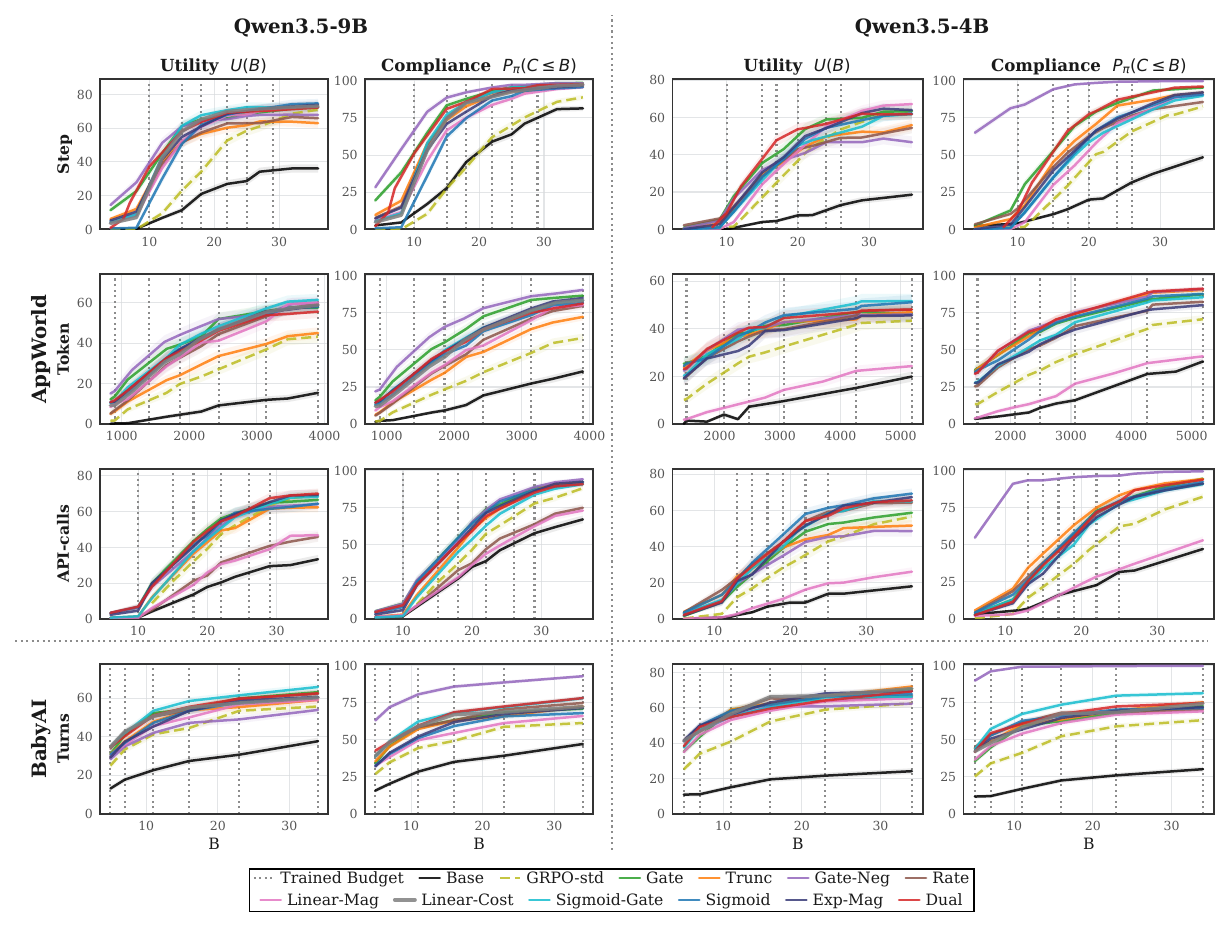}
    \caption{\textbf{In-domain budget utility and compliance.} Budget utility $\mathcal{U}(b)$ and natural compliance $\mathrm{Comp}(b)$ against stated budget $b$ for Qwen3.5-9B and 4B. Rows show AppWorld \texttt{test-normal} under step, token, and API-call budgets, followed by BabyAI under turn budgets. Each policy is trained for one environment and resource type. Vertical dashed lines mark training budgets, with other points unseen during training. Bands show $\pm1$ standard deviation of the binomial rate estimates.}
    \label{fig:appworld-main}
    \vspace{-1em}
\end{figure}

We first test whether a single policy can learn to adapt across dynamically specified budget values, including values unseen during training (\Cref{sec:exp-indomain}). We then compare which reward objectives best teach success within budget and natural compliance (\Cref{sec:exp-objectives}). Finally, we test whether the learned behaviour persists under an AppWorld task-distribution shift and transfers to ASTRA without retraining (\Cref{sec:exp-ood}).

\paragraph{Experimental setup.}
We train 4B and 9B Qwen3.5 models \citep{qwen3.5} from their post-trained checkpoints (\Cref{app:train-grpo}). All trained methods use the same Dr.\ GRPO pipeline \citep{liu2025understanding}, prompts, budget grids, and seed within each setting. They differ in the training return, and \truncname{} additionally stops training rollouts at the budget boundary. We compare against \basename{}, the untrained model, and \grponormalname{}, which is trained with the binary task reward $R$ alone. Like the budget-aware methods, \grponormalname{} receives budget information in its prompt during training and evaluation, but its reward is independent of the budget and its rollouts are not cut off at the budget, unlike \truncname{}. This baseline represents task-reward-only agent training, without a budget-dependent return.

We train and evaluate on AppWorld \citep{trivedi2024appworld}, an interactive coding-agent benchmark where agents complete requests by writing Python over multiple turns. We evaluate on held-out \texttt{test-normal} tasks, the distribution-shifted \texttt{test-challenge} split, and ASTRA \citep{xiu2026astra}, an out-of-domain personal-assistant benchmark. We also evaluate on BabyAI \citep{chevalier-boisvert2019babyai}, a text-based gridworld with tasks distinct from code generation (\Cref{app:datasets}). Across these settings, we consider budgets over \textit{steps} (agent turns), \textit{tool calls} (API calls), and \textit{tokens} (all generated tokens, including reasoning), as defined by the cost map $\kappa$.

At each budget $b$, we measure \emph{budget utility} $\mathcal{U}(b)$, the probability of solving a task within budget, and \emph{compliance} $\mathrm{Comp}(b)$, the probability that the agent terminates within budget when rollouts are not cut off:
\begin{align}
&\mathcal{U}(b) := \mathbb{E}_{x \sim p_{\mathcal{X}_{\mathrm{test}}}, \zeta \sim \pi(\cdot \mid x, b)}\!\left[R(\zeta)\,\mathbf{1}\!\left[\bar{C}(\zeta,b)\leq 1\right]\right],
\\
&\mathrm{Comp}(b) := \mathbb{P}_{x \sim p_{\mathcal{X}_{\mathrm{test}}}, \zeta \sim \pi(\cdot \mid x, b)}\!\left(\bar{C}(\zeta,b)\leq 1\right),
\label{eq:metrics}
\end{align}
where expectations are over held-out tasks $\mathcal{X}_{\mathrm{test}}$. Budget utility counts a success only if the rollout finishes successfully within budget, matching the practical setting in which the agent is cut off at the budget limit. Natural compliance measures whether the rollout finishes within budget regardless of task success, without an external budget cutoff.

\subsection{Agents Learn to Adapt Across Dynamically Specified Budgets}
\label{sec:exp-indomain}

We first test our first question: whether agents can learn to succeed within dynamically specified budgets and adapt their resource use to the task and available allowance.

\paragraph{Budget-shaped returns improve success within budget and compliance.}
Figure~\ref{fig:appworld-main} reports budget utility and natural compliance for \basename{}, \grponormalname{}, and the budget-shaped objectives across both model scales, AppWorld's step, token, and API-call budgets, and BabyAI turn budgets. Despite receiving the same budget information as other methods, \basename{}, the Qwen3.5 checkpoints before our additional training, show poor success within budget and compliance. At a step budget of $B=36$, the 9B and 4B \basename{} models reach only $37\%$ and $19\%$ budget utility, respectively, while the 9B \basename{} model complies on fewer than half of episodes across the token-budget range. \grponormalname{} improves on \basename{} through standard agent post-training, but its budget-independent return still gives weak compliance and lower budget utility than the leading budget-shaped objectives.

The stronger budget-shaped objectives improve both metrics over these baselines, although not every reward improves on \grponormalname{} in every setting. Thus, before looking into differences between the reward functions (which we do in the next section), we can record that budget-shaped post-training, using budget feedback without an external budget-allocation controller, can teach a single agent to use its total and remaining budget to improve success within budget and natural compliance across dynamically specified step, token, and API-call budgets. The BabyAI results show the same overall trends under turn budgets in an environment distinct from AppWorld. At larger budgets, the utility curves of the stronger objectives often converge, indicating that these policies maintain similar task utility once the budget is no longer as binding. No method attains perfect natural compliance across every budget and setting. The objectives yield different trade-offs between success within budget and natural compliance. We compare them in \Cref{sec:exp-objectives}.

\begin{figure}[t]
    \centering
    \includegraphics[width=\linewidth]{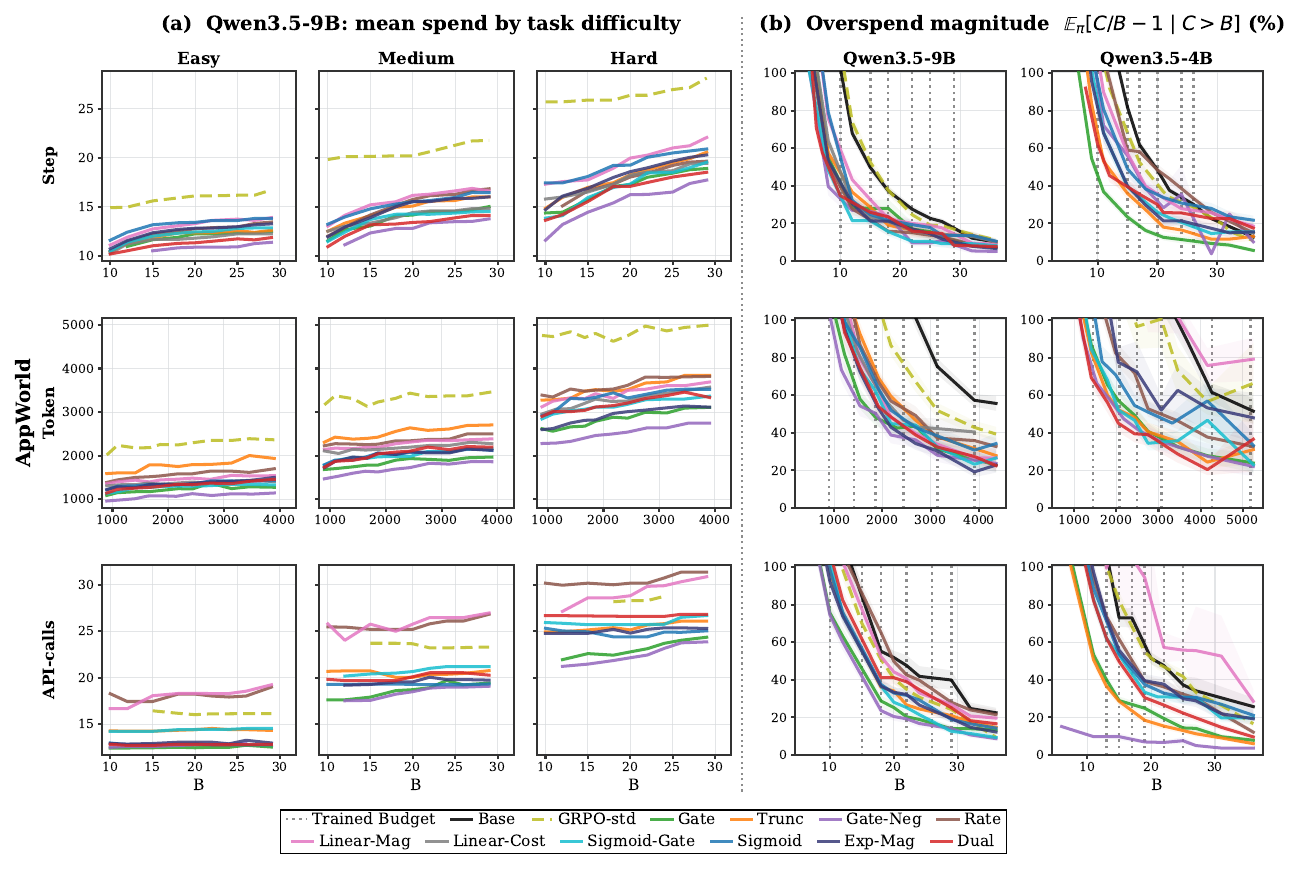}
    \caption{\textbf{Spend by task difficulty and overspend magnitude.} AppWorld \texttt{test-normal} under step, token, and API-call budgets. \textbf{(a)} Mean spend by task difficulty for Qwen3.5-9B. \textbf{(b)} Mean relative overspend $\mathbb{E}_{\pi}[\bar{C}-1 \mid \bar{C}>1]$ for Qwen3.5-9B and 4B, shown as a percentage of the budget. Horizontal axes show the stated budget $B$. Overspend axes are capped at $100\%$.}
    \label{fig:spend}
    \label{fig:excess}
    \vspace{-1em}
\end{figure}

\paragraph{Resource use varies with task difficulty and the stated budget.}
\Cref{fig:spend}(a) shows the mean spend of the 9B AppWorld policies by task difficulty. \grponormalname{} spends comparatively uniformly across stated budgets, despite changing its spend with task difficulty. Across the budget-shaped policies, step and token spend generally increase with the stated budget allowance and with task difficulty, most clearly for hard tasks. Under API-call budgets, spend rises sharply with task difficulty but changes comparatively little with the stated allowance, indicating that call use is driven more strongly by task difficulty. Examples from BabyAI and AppWorld in \Cref{app:qualitative-budget-use} also show explicit budget planning and action batching in the policies' reasoning.

\begin{wraptable}{r}{0.44\linewidth}
  \vspace{-14pt}
  \centering
  \footnotesize
  \setlength{\tabcolsep}{3pt}
  \caption{\textbf{Mean API calls per step by budget.}
  Qwen3.5-9B, restricted to episodes completed within budget on AppWorld \texttt{test-normal} with step budgeting. Bold: highest per budget.}
  \label{tab:told-vs-actual-api-calls-per-step}
  \begin{tabular}{lccc}
  \toprule
   & $B=10$ & $B=15$ & $B=32$\\
  \midrule
  \basename{} & $1.65 \pm 0.56$ & $1.45 \pm 0.53$ & $1.46 \pm 0.40$ \\
  \grponormalname{} & $1.02 \pm 0.05$ & $1.03 \pm 0.16$ & $1.05 \pm 0.17$ \\
  \midrule
  \gatedname{} & $\mathbf{1.93 \pm 0.66}$ & $1.73 \pm 0.50$ & $1.53 \pm 0.39$ \\
  \sigmoidgatename{} & $1.87 \pm 0.69$ & $1.63 \pm 0.45$ & $1.53 \pm 0.36$ \\
  \baponame{} & $1.92 \pm 0.62$ & $\mathbf{1.76 \pm 0.48}$ & $\mathbf{1.61 \pm 0.38}$ \\
  \bottomrule
  \end{tabular}
  \vspace{-14pt}
\end{wraptable}
Table~\ref{tab:told-vs-actual-api-calls-per-step} shows that budget-aware training helps agents use the prompt's guidance to batch API calls (\Cref{code:budget-block-steps}). All methods receive this guidance, but \grponormalname{} uses around one API call per turn across budgets. In contrast, \gatedname{}, \sigmoidgatename{}, and \baponame{} use around $1.9$ calls per turn at $B=10$, compared with around $1.5$ to $1.6$ at $B=32$. These averages include successful and failed rollouts completed within budget. The budget-aware policies batch more calls and increase batching under tighter step budgets. Together with their improved within-budget success and natural compliance, these results suggest that the policies learn to use batching as part of allocating their available steps more effectively.

\paragraph{Budget-aware behaviour generalises beyond the training budgets.}
The dashed vertical lines in \Cref{fig:appworld-main} mark the six budgets used during training. On AppWorld, the step and API-call evaluations include both interpolated budgets between these lines and extrapolated budgets outside the training grid. The stronger budget-aware policies retain improved budget utility and compliance at these unseen values.

\takeawaytext{With total and remaining budget feedback, agents can learn to allocate a shared budget across turns. Budget-shaped training improves success within budget and natural compliance over the base models and over task-reward-only training. Resource use and batching vary with task demands and the available budget, and the gains extend to unseen budget values on AppWorld.}

\subsection{Which Rewards Best Teach Budget-Aware Behaviour?}
\label{sec:exp-objectives}

We now address our second research question by comparing how reward objectives balance success within budget and natural compliance.

\begin{wraptable}{r}{0.42\linewidth}
\vspace{-14pt}
\centering
\footnotesize
\setlength{\tabcolsep}{3pt}
\caption{\textbf{Utility across budgets.} Win and near-best ($\leq$2pp) rates and mean regret against the per-budget best method, averaged equally over the eight AppWorld and BabyAI experiments (\Cref{app:ranking-robustness}).}
\label{tab:main-winrate-pooled}
\begin{tabular}{lccc}
\toprule
Method & \makecell{Win\\(\%) $\uparrow$} & \makecell{Near-best\\(\%) $\uparrow$} & \makecell{Regret\\(pp) $\downarrow$} \\
\midrule
\magname{}          & 3.1  & 6.5  & 14.01 \\
\truncname{}        & 6.6  & 20.5 & 7.56 \\
\ratename{}         & 3.1  & 26.0 & 7.46 \\
\gatenegname{}      & 16.1 & 26.0 & 6.00 \\
\expmagname{}       & 7.8  & 22.9 & 4.32 \\
\sigmoidname{}      & 14.0 & 39.2 & 4.31 \\
\sigmoidgatename{}  & \textbf{20.1} & 36.9 & 3.82 \\
\costname{} & 11.6 & 39.3 & 3.75 \\
\gatedname{}        & 12.4 & \textbf{39.8} & 3.33 \\
\baponame{}         & 10.8 & 38.8 & \textbf{2.92} \\
\bottomrule
\end{tabular}
\vspace{-6pt}
\end{wraptable}
\paragraph{Utility rankings of budget-shaped rewards.}
\Cref{tab:main-winrate-pooled} aggregates budget utility across the eight AppWorld and BabyAI experiments. Near-best rate measures how often a method is within two percentage points of the best budget utility at each budget, while mean regret is the average gap to the best observed budget utility at each budget, measured in percentage points. \sigmoidgatename{} most often attains the highest utility at an individual budget, with a $20.1\%$ win rate. \gatedname{} has the highest near-best rate at $39.8\%$. \baponame{} has the lowest observed mean regret at $2.92$ percentage points, followed by \gatedname{} at $3.33$, \costname{} at $3.75$, and \sigmoidgatename{} at $3.82$. Thus, \sigmoidgatename{} wins most often at an individual budget and \baponame{} has the smallest observed average loss to the best method across the evaluation. \gatedname{} performs surprisingly well for a simple objective that rewards only within-budget success and adds no explicit overspend penalty.

\paragraph{Allowing training rollouts beyond the budget improves pooled utility.}
\Cref{tab:main-winrate-pooled} shows that \gatedname{} surpasses \truncname{} on all three pooled utility metrics, with a higher win rate and near-best rate and a lower mean regret. \gatedname{} permits an overspending rollout to finish and assigns it zero return. Tokens generated after the budget boundary remain in the rollout and contribute to training. When the same group contains a positive-return rollout, the overspending rollout receives a negative advantage, providing a learning signal on its actions, including those beyond the budget. \truncname{} instead ends the rollout at the budget boundary, removing these post-budget tokens from training. The comparison supports retaining overspending trajectories even when they receive no success reward. Note, however, that \truncname{} has a higher throughput due to not having to generate the additional over budget steps.

\paragraph{Penalty design affects success, compliance, and their trade-off.}
\gatenegname{} assigns every overspent trajectory a reward of $-1$, ranking an in-budget failure above any overspent result. In \Cref{fig:appworld-main}, \gatenegname{} generally attains high natural compliance, but its mean regret is $6.00$ percentage points, compared with $3.33$ for \gatedname{} in \Cref{tab:main-winrate-pooled}. The strong overspend penalty may make the policy more hesitant near the budget boundary, sacrificing some success within budget to avoid a violation.

\ratename{} alone is less competitive than \baponame{}, with mean regrets of $7.46$ and $2.92$ percentage points, respectively. This is consistent with a learned price on the overspend event being more useful when combined with information about the overspend size. \costname{} and \sigmoidname{}, which apply cost pressure before the budget boundary, have mean regrets of $3.75$ and $4.31$ percentage points, respectively. These results suggest that early pressure to finish does not necessarily improve within-budget utility.

\begin{figure}[t]
    \centering
    \includegraphics[width=0.85\linewidth]{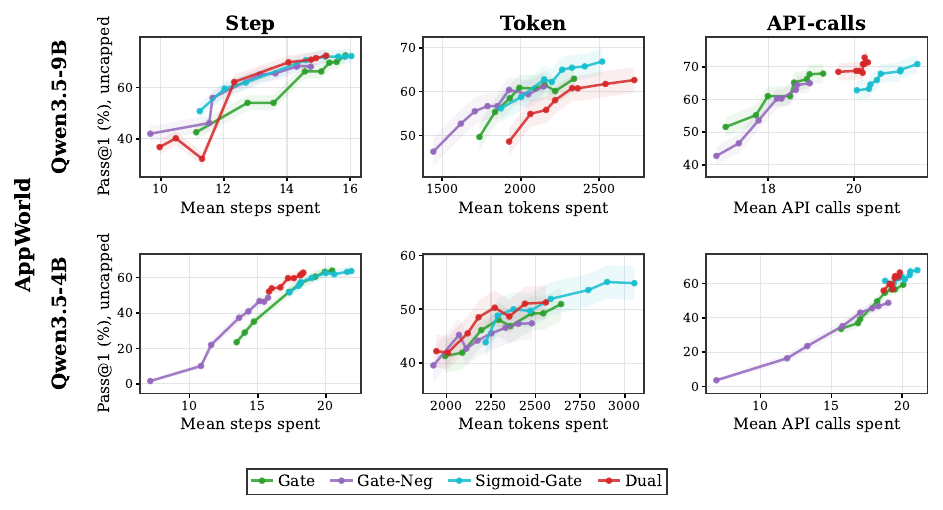}
    \captionsetup{skip=2pt}
    \caption{\textbf{Raw task success against spend.} AppWorld \texttt{test-normal} pass@1 without a budget cutoff against mean spend for \gatedname{}, \gatenegname{}, \sigmoidgatename{}, and \baponame{}. Each point corresponds to a stated budget. Policies retain budget feedback, but successes count regardless of compliance. Bands show $\pm1$ standard deviation of the binomial rate estimates.}
    \label{fig:pass-at-1}
\end{figure}
\paragraph{Small overruns without magnitude penalties.}
Even a policy with high natural compliance can substantially overspend on the episodes where it violates the budget. We therefore also examine conditional excess spend in \Cref{fig:excess}(b), that is, in rollouts that do overspend, how much do they overspend in percent compared to the prompted budget. \gatedname{} and \gatenegname{} generally have comparatively small conditional overruns, despite neither applying a penalty that grows with overspend magnitude. Explicit magnitude penalties are therefore not necessary to obtain comparatively small overruns in these experiments.

\paragraph{Raw task success generally increases with spend.}
This paper so far focussed on the evaluation setup where the agent is cut off at the budget in each task. We may, however, also be interested in the tradeoff between how much compute is used overall versus how many tasks are solved. \Cref{fig:pass-at-1} reports pass@1 against mean spend, counting successes regardless of whether the rollout finishes within budget. At both model scales, greater spending is generally associated with higher overall success across steps, tokens, and API calls. This complements the earlier evidence of budget-responsive resource use by showing that spending differences are also accompanied by differences in overall task success.

\takeawaytext{Reward objectives produce different trade-offs between success within budget and natural compliance. \baponame{} is closest to the per-budget optimal method on average, \sigmoidgatename{} most often achieves the highest within-budget success at an individual budget, and \gatenegname{} generally achieves the highest natural compliance. Allowing training rollouts to continue beyond the budget can improve success within budget. The simple \gatedname{} objective, which rewards only success within budget, is often surprisingly competitive with more complex methods.
}

\FloatBarrier
\subsection{Budget-Aware Behaviour Generalises Beyond the Training Distribution}
\label{sec:exp-ood}

\begin{figure}[t]
    \centering
    \includegraphics[width=0.98\linewidth]{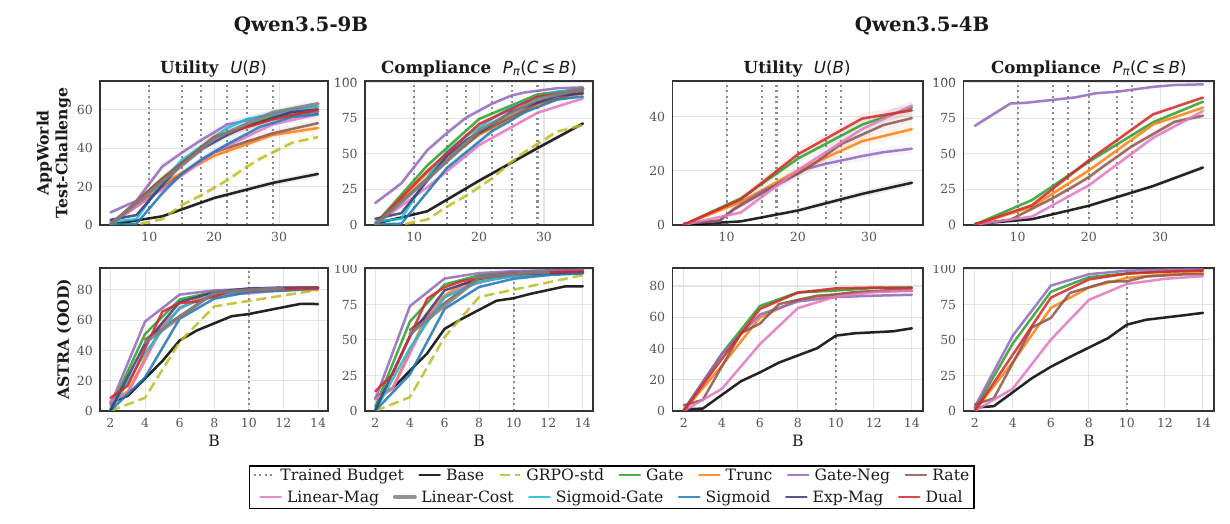}
    \caption{\textbf{Distribution shift.} AppWorld-trained policies evaluated without retraining on AppWorld \texttt{test-challenge} (top) and ASTRA (bottom) under step budgets. Curves show budget utility $\mathcal{U}(b)$ and natural compliance $\mathrm{Comp}(b)$ against stated budget $b$. Bands show $\pm1$ standard deviation of the binomial rate estimates.}
    \label{fig:ood-eval}
\end{figure}

\paragraph{Budget-aware behaviour persists on shifted AppWorld tasks.}
We next test whether policies trained to condition on the budget remain responsive to it under distribution shift. \Cref{fig:ood-eval} (top) evaluates the AppWorld-trained policies on \texttt{test-challenge}, which includes unseen apps and more difficult tasks. As on \texttt{test-normal}, the stronger budget-shaped objectives generally achieve higher budget utility and natural compliance than \basename{} and \grponormalname{} at both model scales. Utility rises with the stated step budget and, for the 9B model, continues to increase beyond the largest training budget of $29$. These results suggest that both the learned budgeting behaviour and the gains over budget-independent training methods extend to shifted tasks. \gatenegname{} once again generally achieves the highest compliance, while utility rankings vary with the budget and model scale.

\paragraph{Budget-aware behaviour transfers zero-shot to ASTRA.}
We also test zero-shot generalization to a new environment. \Cref{fig:ood-eval} (bottom) evaluates the same policies on ASTRA without retraining. As on \texttt{test-challenge}, the stronger budget-shaped objectives generally retain higher budget utility and natural compliance than \basename{} and \grponormalname{}. At both model scales, utility increases over tighter step budgets before flattening at looser budgets, while natural compliance also increases with the allowance. \gatenegname{} remains among the most compliant methods, and several objectives achieve similar utility at looser budgets. These results suggest that budget responsiveness and the improvements over the baselines transfer to ASTRA's distinct personal-assistant tasks and tool domain, using the same Python-REPL action interface (\Cref{app:data-astra}).

\FloatBarrier
\takeawaytext{Budget-aware policies remain responsive to stated budgets under distribution shift and zero-shot transfer to ASTRA. \gatenegname{} remains the best choice when prioritising natural budget adherence without requiring external enforcement.}

\section{Conclusion}
\label{sec:conclusion}
Simply prompting an agent with an inference budget is not enough to make it adhere to that budget or succeed within it. Without consequences for overspending during training, agents frequently exceed the stated allowance. Budget-aware reward objectives provide a simple remedy, teaching a single policy to adapt its resource use across tasks and budget values. We show that this behaviour extends to unseen budgets, persists under a shifted task distribution, and transfers zero-shot to a separate environment. The choice of objective determines the resulting trade-off. \baponame{} is most consistently close to the best method across budgets, \sigmoidgatename{} most often yields the highest within-budget success, and \gatenegname{} yields the strongest natural compliance. All three are competitive and require no external budget-allocation controller. We encourage practitioners to experiment with these objectives to select the balance between task success and natural budget adherence that best fits their application.

All tool-use experiments in this work are conducted in research-only simulated environments. The APIs, tasks, datasets, verifier infrastructure, and training procedures described here do not represent or imply any deployed Apple product, production system, or product roadmap.

\applefootnote{\textcolor{textgray}{\sffamily Apple and the Apple logo are trademarks of Apple Inc., registered in the U.S. and other countries and regions.}}

\setlength{\bibsep}{3pt plus 1pt}
\bibliographystyle{plainnat}
\bibliography{iclr2026_conference}

\clearpage
\beginappendix
\crefalias{section}{appendix}
\crefalias{subsection}{subappendix}
\crefalias{subsubsection}{subsubappendix}

\startcontents[appendices]
\printcontents[appendices]{}{1}{\section*{Appendix Contents}\setcounter{tocdepth}{2}}
\clearpage

\section{Notation}
\label{app:notation}

\Cref{tab:notation} acts as a reference for the notation used throughout the paper, grouped by where they are introduced.

\begin{table}[H]
\centering
\small
\begin{tabularx}{\textwidth}{l X}
\toprule
\multicolumn{2}{l}{\textbf{Task and environment (\Cref{sec:prelim,sec:bapomdp})}} \\
\midrule
$\mathcal{S}, \tilde{\mathcal{S}}, \mathcal{A}, \Omega, \tilde{\Omega}$ & Task and budget-augmented state spaces, action space, and observation spaces \\
$\mathcal{T}, \tilde{\mathcal{T}}, \mathcal{O}, \tilde{\mathcal{O}}$ & Task and budget-augmented transition and observation kernels \\
$\mathcal{M}, \tilde{\mathcal{M}}$ & Task POMDP and budget-augmented POMDP \\
$\gamma, \mu_0$ & Discount factor and initial state distribution \\
$x, p_{\mathcal{X}}$ & Task and task distribution over $\mathcal{X}$ \\
$h_t, \mathcal{H}$ & Observation history at step $t$, and set of all histories \\
$\pi, \Pi$ & Policy, and set of policies \\
$\kappa$ & Realised transition cost, abbreviated as $\kappa(a_t)$ \\
$a_\perp$ & \texttt{complete\_task} action that ends an episode \\
$T_{\max}, \tau, \tau_{\mathrm{env}}$ & Truncation horizon, episode stopping time, and environment termination time \\
$\zeta$ & Completed trajectory \\
$R(\zeta)$ & Binary terminal reward \\
\midrule
\multicolumn{2}{l}{\textbf{Budgets and cost (\Cref{sec:bapomdp} and \Cref{sec:perbudget})}} \\
\midrule
$B_0, b_t$ & Episode budget, and remaining budget at step $t$ \\
$\mathcal{B}, M$ & Finite set of training budgets, and its size $M = |\mathcal{B}|$ \\
$p_B$ & Budget prior over $\mathcal{B}$ \\
$C(\zeta)$ & Raw trajectory spend, e.g.\ steps, tool calls, or tokens \\
$\bar{C}(\zeta, B_0)$ & Normalised trajectory cost, $C(\zeta) / B_0$ \\
$\delta$ & Permitted overspend rate \\
\midrule
\multicolumn{2}{l}{\textbf{Objectives and rewards (\Cref{sec:shaped} and \Cref{app:theory})}} \\
\midrule
$P_{\mathrm{DAPO}}$ & Soft-overlong penalty, adapted from DAPO \\
$\alpha$ & Coefficient of $P_{\mathrm{DAPO}}$, fixed at $1/2$ \\
$\lambda_b$ & Lagrange multiplier at budget $b$ \\
$\Lambda_b$ & Effective price at budget $b$, $\Lambda_b = \lambda_b / p_B(b)$ \\
$\mathcal{L}_{\mathrm{rate}}, \mathcal{L}_{\mathrm{dual}}$ & Lagrangians of \ratename{} and \baponame{}, with $\mathcal{L}$ abbreviating $\mathcal{L}_{\mathrm{dual}}$ in the derivations \\
$\tilde{R}(\zeta, B_0)$ & Shaped return \\
$\tilde{R}_{\mathrm{gate}}, \tilde{R}_{\mathrm{trunc}}$ & Gated and truncated within-budget success rewards \\
$\tilde{R}_{\mathrm{DAPO}}, \tilde{R}_{\mathrm{rate}}$ & Rewards with only the overspend magnitude or only the overspend rate penalty \\
$J(\pi), \tilde{J}(\pi)$ & Success objective, and its shaped counterpart \\
$g_b(\pi)$ & Constraint functional at budget $b$ \\
\midrule
\multicolumn{2}{l}{\textbf{Policy optimisation (\Cref{app:grpo,app:updates})}} \\
\midrule
$\theta, \pi_\theta$ & Policy parameters, and the policy they induce \\
$K, G$ & Number of task-budget groups per iteration, and rollouts per group \\
$A^{(g,i)}$ & Dr.\ GRPO advantage of rollout $i$ in group $g$ \\
$\sigma(\cdot), \sigma^{(g)}$ & Logistic sigmoid in the reward definitions, and group reward standard deviation in GRPO \\
$\epsilon$ & PPO clip range \\
$r_{t,l}^{(g,i)}(\theta)$ & Token-level importance ratio \\
$\beta$ & Dual step size \\
$\Lambda_{\max}$ & Cap on the effective price \\
$\hat{p}_b, \mathcal{I}_b$ & Batch overspend rate at budget $b$, and the rollouts it is estimated from \\
\midrule
\multicolumn{2}{l}{\textbf{Evaluation (\Cref{sec:experiments})}} \\
\midrule
$\mathcal{U}(b), \mathrm{Comp}(b)$ & Budget utility and compliance at budget $b$ \\
$\mathcal{X}_{\mathrm{test}}, p_{\mathcal{X}_{\mathrm{test}}}$ & Held-out evaluation task set and its distribution \\
$\bar{C}(\zeta,b)$ & Normalised trajectory cost at budget $b$ \\
\bottomrule
\end{tabularx}
\caption{\textbf{Notation.} Symbols used throughout the main paper and the appendix.}
\label{tab:notation}
\end{table}

\section{Limitations and Future Work}
\label{sec:limitations}

Each policy currently manages one resource type at a time. Extending the framework to jointly manage several resources would allow an agent to balance constraints such as tokens, tool calls, and environment steps within the same episode. Although the remaining budget is updated after every turn, the total budget and resource definition are fixed at the start of the episode. Future work could consider budgets that are revised during execution or costs that change with the environment state. The framework could also incorporate stochastic costs such as latency or monetary spend and be combined with inference-time allocation methods that build on the budget awareness already internalised by the policy.

Our evidence covers two scales of one model family and simulated environments. Most method comparisons use one training seed; the additional seeds cover two objectives in one AppWorld setting and do not establish a reliable ordering among all leading methods. Unseen-budget evaluation covers AppWorld interpolation for all three resources and extrapolation for steps and API calls; BabyAI is evaluated only at trained budgets. ASTRA transfer uses step budgets and an action interface aligned with AppWorld.

Natural compliance is imperfect, and the capped multiplier updates do not guarantee the target violation rate. A strict deployment budget therefore still requires a cutoff. Our prompts provide remaining-budget feedback and resource-specific advice, including instructions to batch actions. Without ablations of these prompt components, the experiments establish the behaviour of the combined prompting and training procedure, rather than the separate contribution of each component.

\section{Budget-Conditioned Rewards in Full}
\label{app:rewards}

This appendix defines every reward of \Cref{fig:reward-functions} and derives the per-budget price of \ratename{} and \baponame{}. All rewards are written using normalised cost $\bar{C} = \bar{C}(\zeta, B_0)$ of \Cref{eq:normcost}, so that the budget sits at $\bar{C} = 1$ and allows a generalised application across budget types.

\paragraph{Hard gates and fixed penalties.}
\begin{itemize}[leftmargin=1em, topsep=2pt, itemsep=5pt, partopsep=0pt, parsep=0pt]
\item \truncname{} enforces the budget in the environment. It halts the episode before executing any action that would overdraw the budget, at $\tau_{B_0} := \min\big(\tau,\ \inf\{ t \geq 1 : b_t - \kappa(a_t) < 0 \} - 1\big)$, and scores the truncated trajectory by the task reward alone, $\tilde{R}_{\mathrm{trunc}}(\zeta, B_0) := R(\zeta_{\leq \tau_{B_0}})$. Every training trajectory then complies by construction, but the policy receives no training on continuations beyond the budget boundary. Compliance during training is therefore not evidence that it has learned to stop without a cutoff.
\item \gatedname{} lets the rollout continue past the budget but withholds the task reward from any overspending trajectory, $\tilde{R}_{\mathrm{gate}}(\zeta, B_0) := R\,\mathbf{1}[\bar{C} \leq 1]$.
\item \gatenegname{} \citep{team2026kimi} extends \gatedname{} with a fixed penalty for overspending, $R\,\mathbf{1}[\bar{C} \leq 1] - \mathbf{1}[\bar{C} > 1]$.
\item \costname{} \citep{deepseek2026v41flash} penalises cost linearly from the first action, $R - \min(1, \bar{C}/2)$, so it applies pressure before the budget is reached.
\end{itemize}

\paragraph{Continuous cost shaping.}
\sigmoidgatename{} and \sigmoidname{} use the logistic sigmoid $\sigma(x) := (1 + e^{-x})^{-1}$ to supply continuous cost feedback. They differ in whether that feedback starts only after the budget or also acts within it.
\begin{itemize}[leftmargin=1em, topsep=2pt, itemsep=5pt, partopsep=0pt, parsep=0pt]
\item \sigmoidgatename{} has return $R + 1 - 2\sigma\big(2(\bar{C}-1)_+\big)$. Its additive penalty is zero for $\bar{C}\leq1$ and increases continuously towards $1$ beyond the budget. Thus, overspending successes approach return $0$ and overspending failures approach $-1$. Despite its name, this is not a multiplicative smoothing of \gatedname{}'s success gate; the positive-part operation also introduces a kink at the budget.
\item \sigmoidname{} smooths the overspend indicator $\mathbf{1}[\bar{C} > 1]$ that \ratename{} and \baponame{} apply as a hard step, $R - \sigma\big(2(\bar{C}-1)\big)$. Unlike \sigmoidgatename{}, this penalty also acts within budget, with a slope that varies with cost.
\end{itemize}

\paragraph{Magnitude shaping.}
We adapt the DAPO shaping \Cref{eq:overlong} to the normalised cost so that one form covers every budget in every cost unit,
\begin{equation}
P_{\mathrm{DAPO}}(\zeta, B_0) :=
\alpha \operatorname{clip}\!\big(\, \bar{C}(\zeta, B_0) - 1,\ 0,\ 1 \,\big) ,
\label{eq:softoverlong}
\end{equation}
where $\alpha := 1/2$ is set by DAPO's reward scale and is not tuned (derived in \Cref{app:dapo}). Its reward $\tilde{R}_{\mathrm{DAPO}} := R - P_{\mathrm{DAPO}}$ extends DAPO from one cap fixed before training to a budget sampled each episode. Its fixed coefficient penalises how far a policy overspends but not how often. \expmagname{} instead decays a success exponentially in the overspend, $R\, e^{-2(\bar{C} - 1)_+}$, and leaves every failure at $0$.

\paragraph{A per-budget constrained objective.}
\label{sec:perbudget}
A cost penalty sets a single trade-off between success and spending, but does not directly control how often the policy overspends. To define the \ratename{} and \baponame{} objectives, we instead maximise success subject to a separate constraint on the probability of overspending at each budget. This gives the constrained objective
\begin{equation}
\max_{\pi \in \Pi}\ \ \mathbb{E}_{\substack{x \sim p_{\mathcal{X}},\ B_0 \sim p_B \\ \zeta \sim \pi(\cdot \mid x, B_0)}}\!\big[\, R(\zeta) \,\big]
\quad \text{s.t. }\ \ \mathbb{P}_{\substack{x \sim p_{\mathcal{X}} \\ \zeta \sim \pi(\cdot \mid x, b)}}\!\big(\, \bar{C}(\zeta, b) > 1 \,\big) \leq \delta,
\quad \forall\, b \in \mathcal{B},
\label{eq:constrained}
\end{equation}
where $\delta \in (0, 1)$ is the permitted overspend rate. The objective averages over tasks and budgets together, whereas each constraint conditions on the budget and averages over tasks alone, so it is the overspend rate a user would measure at budget $b$ across the distribution of tasks that they wish to have completed. Slack may therefore move between tasks but never between budgets.

Subtracting $P_{\mathrm{DAPO}}$ from the objective of \Cref{eq:constrained}, with the constraints unchanged, gives our \textit{\baponame{} training objective},
{\small
\begin{equation}
\max_{\pi \in \Pi}\ \ \mathbb{E}_{\substack{x \sim p_{\mathcal{X}},\ B_0 \sim p_B \\ \zeta \sim \pi(\cdot \mid x, B_0)}}\!\big[\, R(\zeta) - P_{\mathrm{DAPO}}(\zeta, B_0) \,\big]
\quad \text{s.t. }\ \ \mathbb{P}_{\substack{x \sim p_{\mathcal{X}} \\ \zeta \sim \pi(\cdot \mid x, b)}}\!\big(\, \bar{C}(\zeta, b) > 1 \,\big) \leq \delta,
\quad \forall\, b \in \mathcal{B}.
\label{eq:bapo}
\end{equation}}

\paragraph{Lagrangian relaxation.}
Introducing a non-negative multiplier $\lambda_b$ for each budget-specific constraint gives the Lagrangian for \ratename{}:
{\small
\begin{equation}
\mathcal{L}_{\mathrm{rate}}\big(\pi, \{\lambda_b\}\big)
= \mathbb{E}_{\substack{x \sim p_{\mathcal{X}},\ B_0 \sim p_B \\ \zeta \sim \pi(\cdot \mid x, B_0)}}\!\big[\, R(\zeta) - \Lambda_{B_0}\, \mathbf{1}[\bar{C}(\zeta, B_0) > 1] \,\big]
+ \delta \sum_{b \in \mathcal{B}} \lambda_b .
\label{eq:lagrangian-rate}
\end{equation}}%
Adding the DAPO magnitude penalty gives the Lagrangian for \baponame{}:
{\small
\begin{equation}
\mathcal{L}_{\mathrm{dual}}\big(\pi, \{\lambda_b\}\big)
= \mathbb{E}_{\substack{x \sim p_{\mathcal{X}},\ B_0 \sim p_B \\ \zeta \sim \pi(\cdot \mid x, B_0)}}\!\big[\, R(\zeta) - P_{\mathrm{DAPO}}(\zeta, B_0) - \Lambda_{B_0}\, \mathbf{1}[\bar{C}(\zeta, B_0) > 1] \,\big]
+ \delta \sum_{b \in \mathcal{B}} \lambda_b .
\label{eq:lagrangian}
\end{equation}}%
\noindent Here $\Lambda_b := \lambda_b / p_B(b)$ is the effective price of overspending at budget $b$; the factor $1/p_B(b)$ corrects for how often that budget is sampled. The final term is constant in $\pi$, leaving the per-trajectory shaped return used by each method. In particular, $\tilde{R}_{\mathrm{rate}} := R - \Lambda_{B_0}\mathbf{1}[\bar{C}>1]$, and \baponame{} additionally subtracts $P_{\mathrm{DAPO}}$. Both methods update the multipliers in response to the overspend rate at each budget (\Cref{app:optimisation}). The rate penalty targets how often the policy overspends, while $P_{\mathrm{DAPO}}$ in \baponame{} additionally distinguishes overspending trajectories by how far they exceed the budget (\Cref{app:roles}).

\section{Adapting DAPO to Multi-Budget Settings}
\label{app:dapo}
We give DAPO's original soft-overlong penalty and show how we adapt it from a single fixed token cap to a family of budgets.

DAPO \citep{yu2025dapo} introduces a soft-overlong penalty for truncated responses in long-chain-of-thought reinforcement learning, evaluated on mathematical reasoning tasks. The penalty increases with response length within a soft window near the maximum generation length. Given a maximum response length $L_{\max} \in \mathbb{Z}_{>0}$ and a soft window $L_{\mathrm{s}} \in \mathbb{Z}_{>0}$ with $L_{\mathrm{s}} \leq L_{\max}$, it shapes the reward according to the response length $L(o) \in \mathbb{Z}_{\geq 0}$ as
\begin{equation}
R_{\mathrm{len}}(o) =
\begin{cases}
0, & L(o) \leq L_{\max} - L_{\mathrm{s}}, \\[4pt]
\dfrac{(L_{\max} - L_{\mathrm{s}}) - L(o)}{L_{\mathrm{s}}}, & L_{\max} - L_{\mathrm{s}} < L(o) \leq L_{\max}, \\[4pt]
-1, & L(o) > L_{\max}.
\end{cases}
\label{eq:overlong}
\end{equation}
The penalty is zero up to the soft threshold $L_{\max} - L_{\mathrm{s}}$, decreases linearly to $-1$ over a window of width $L_{\mathrm{s}}$, and remains at $-1$ beyond $L_{\max}$. It therefore supplies a continuous penalty based on overspend magnitude.

This construction was designed for a single token-length cap fixed before training. Our setting instead requires one policy to work across a family of budgets $B_0$, with the budget varying between episodes, and with budgeted quantities other than tokens, such as steps or tool calls. We adapt DAPO's penalty to this setting by expressing spend relative to the budget of the current episode.

Concretely, set $L_{\mathrm{s}} = B_0$ and $L_{\max} = 2B_0$ in \Cref{eq:overlong}. The soft threshold $L_{\max} - L_{\mathrm{s}}$ is then exactly $B_0$, and the middle branch becomes
\begin{equation}
\frac{(2B_0-B_0)-L(o)}{B_0}
=
\frac{B_0-L(o)}{B_0}
=
-\left(\frac{L(o)}{B_0}-1\right).
\label{eq:dapo-substitution}
\end{equation}
We therefore apply the linear ramp to the normalised trajectory cost $\bar{C}(\zeta,B_0)$ of \Cref{eq:normcost} instead of to the raw token count $L(o)$ against one fixed cap. Reversing the sign so that it is written as a positive penalty to be subtracted from the return gives the same ramp in normalised-cost units,
$\operatorname{clip}\!\left(\bar{C}(\zeta,B_0)-1,\,0,\,1\right).$
The cost map $\kappa$ determines the quantity being budgeted, while normalisation by $B_0$ makes the soft threshold $\bar{C}(\zeta,B_0)=1$ for every budget. The same DAPO penalty therefore applies to any budget and to any cost unit.

It remains to match the scale of DAPO's original reward. DAPO pairs a task reward $R_{\mathrm{DAPO}} \in \{-1,+1\}$ with a penalty $R_{\mathrm{len}}\in[-1,0]$, whereas our terminal reward $R(\zeta) \in \{0,1\}$. The map $x \mapsto (x+1)/2$ takes DAPO's task-reward scale to ours. Applying the same map to DAPO's shaped return gives
\begin{equation}
\frac{R_{\mathrm{DAPO}} + R_{\mathrm{len}} + 1}{2}
= \underbrace{\frac{R_{\mathrm{DAPO}} + 1}{2}}_{R(\zeta)}
\; + \; \tfrac{1}{2} R_{\mathrm{len}} ,
\label{eq:rescale}
\end{equation}
so preserving DAPO's relative weighting between task success and the soft-overlong penalty introduces a factor of $\tfrac{1}{2}$. Combining this rescaling with the normalised expression above gives our penalty
\begin{equation}
P_{\mathrm{DAPO}}(\zeta,B_0)
=
\tfrac{1}{2}\operatorname{clip}\!\left(\bar{C}(\zeta,B_0)-1,\,0,\,1\right),
\label{eq:dapo-adaptation}
\end{equation}
which is \Cref{eq:softoverlong} with $\alpha=\tfrac{1}{2}$. We therefore set $\alpha=0.5$ throughout (\Cref{app:train-budget}). In raw spend, $C$, the penalty has slope $1/(2B_0)$ within the linear overspending region, so a fixed absolute overspend carries a larger penalty under a tighter budget than under a looser budget.

The rescaling preserves the relative weighting of task success and the length penalty. The Dr.\ GRPO advantage centres each trajectory's return within its group (\Cref{app:updates}), so an affine transformation $\tilde{R} \mapsto a\tilde{R}+c$ with $a>0$ removes the additive constant under centring and multiplies every advantage by $a$. For the unfiltered surrogate, this preserves the advantage signs and relative magnitudes and rescales its gradient by a constant.

\section{Optimising Budget-Conditioned Objectives}
\label{app:theory}

This appendix describes the optimisation used in \Cref{sec:method}. \Cref{app:setup} fixes the policy class and constraint functionals, \Cref{app:rearrangement} verifies the rearrangement of the Lagrangian \Cref{eq:lagrangian}, \Cref{app:gradients} derives the policy and multiplier updates of \Cref{app:optimisation} from it, and \Cref{app:roles} explains the distinct signals supplied by \baponame{}'s two penalties. \Cref{app:grpo,app:updates} give the shared Dr.\ GRPO policy update and the additional multiplier updates used by \ratename{} and \baponame{}. \Cref{app:train-algo} gives a general training algorithm for fixed reward objectives, followed by the shared algorithm for \ratename{} and \baponame{}.

\subsection{Background}
Constrained MDPs (cMDPs) typically bound expected cumulative cost for a single fixed threshold \citep{altman1999constrained, achiam2017cpo}, often using Lagrangian methods for optimisation \citep{stooke2020responsive}. Budgeted MDPs \citep{carrara2019budgeted} extend cMDPs to families of budgets by conditioning value functions on the remaining allowance, but still constrain \emph{expected} discounted cost return via fitted Q-iteration. This does not directly express budget compliance at the level of individual requests, where an expected-cost constraint can hide frequent individual violations. Risk-aware and chance-constrained formulations instead bound the probability of undesirable events \citep{geibel2005risk, haskell2015convex}, and \citet{chow2017risk} studies trajectory-level probabilistic constraints under Lagrangian relaxation.

\subsection{Setup}
\label{app:setup}

Throughout we work with the trajectory law induced by a policy $\pi$ on the budget-augmented POMDP $\tilde{\mathcal{M}}$ of \Cref{sec:bapomdp}. We write $\mathbb{P}_\pi$ for that law, i.e.\ $\mathbb{P}_\pi(\cdot) = \mathbb{P}_{x \sim p_{\mathcal{X}},\ B_0 \sim p_B,\ \zeta \sim \pi(\cdot \mid x, B_0)}(\cdot)$ and likewise $\mathbb{E}_\pi$ for expectation under it, matching the sampling made explicit in \Cref{eq:constrained,eq:lagrangian}. We include the task draw in this law rather than writing it separately because every quantity below averages over tasks. Conditioning on a fixed budget, written $\mathbb{P}_\pi(\cdot \mid B_0 = b)$ below, still draws $x \sim p_{\mathcal{X}}$ and then rolls out $\zeta$ under $\pi(\cdot \mid x,b)$. We define the objective and per-budget constraint functionals
\begin{equation}
J(\pi) := \mathbb{E}_\pi\!\left[\, R(\zeta) \,\right],
\qquad
g_b(\pi) := \mathbb{P}_\pi\!\left(\, \bar{C}(\zeta, B_0) > 1 \,\middle|\, B_0 = b \,\right) - \delta,
\qquad b \in \mathcal{B},
\label{eq:functionals}
\end{equation}
so that the constrained objective \Cref{eq:constrained} reads $\max_{\pi \in \Pi} J(\pi)$ subject to $g_b(\pi) \leq 0$ for every $b \in \mathcal{B}$. The conditioning in \Cref{eq:functionals} is well-defined because $p_B$ has full support on the finite set $\mathcal{B}$. \Cref{eq:bapo} keeps these constraints and shapes the objective,
\begin{equation}
\tilde{J}(\pi) := \mathbb{E}_\pi\!\left[\, R(\zeta) - P_{\mathrm{DAPO}}(\zeta, B_0) \,\right],
\label{eq:shaped_functional}
\end{equation}
so that it reads $\max_{\pi \in \Pi} \tilde{J}(\pi)$ subject to the same $g_b(\pi) \leq 0$. Its Lagrangian is $\mathcal{L}_{\mathrm{dual}}$ in \Cref{eq:lagrangian}, which we abbreviate as $\mathcal{L}$ below. Results below are stated for $\tilde{J}$, and setting $\alpha = 0$ recovers $J$ and \Cref{eq:constrained}.

\subsection{Rearrangement of the Lagrangian}
\label{app:rearrangement}

Relaxing \Cref{eq:bapo} prices each budget with its own multiplier, through the penalty
\[
\sum_{b \in \mathcal{B}} \lambda_b \big( \mathbb{P}_\pi( \bar{C}(\zeta, B_0) > 1 \mid B_0 = b ) - \delta \big).
\]
Folding that sum into a single expectation produces the compact Lagrangian \Cref{eq:lagrangian} and its per-trajectory integrand, the shaped return of \baponame{} (\Cref{fig:reward-functions}). Writing each overspend probability as an expectation and splitting the outer expectation over the prior $B_0 \sim p_B$,
\begin{align}
\sum_{b \in \mathcal{B}} \lambda_b\, \mathbb{P}_\pi\!\left( \bar{C}(\zeta, B_0) > 1 \,\middle|\, B_0 = b \right)
&= \sum_{b \in \mathcal{B}} \lambda_b\, \mathbb{E}_\pi\!\left[\, \mathbf{1}[\bar{C}(\zeta, B_0) > 1] \,\middle|\, B_0 = b \,\right] \nonumber \\
&= \sum_{b \in \mathcal{B}} p_B(b)\, \frac{\lambda_b}{p_B(b)}\, \mathbb{E}_\pi\!\left[\, \mathbf{1}[\bar{C}(\zeta, B_0) > 1] \,\middle|\, B_0 = b \,\right] \nonumber \\
&= \mathbb{E}_\pi\!\left[\, \frac{\lambda_{B_0}}{p_B(B_0)}\, \mathbf{1}[\bar{C}(\zeta, B_0) > 1] \,\right],
\label{eq:rearrangement}
\end{align}
where the second line divides and multiplies by $p_B(b) > 0$ and the third recognises the resulting sum as an expectation over $B_0 \sim p_B$. The importance weight $1 / p_B(B_0)$ therefore corrects for the frequency with which each budget is sampled, so that a budget contributes to the penalty in proportion to $\lambda_b$ alone and not to how often it is drawn. The remaining term $\delta \sum_b \lambda_b$ is constant in $\pi$ and plays no part in the policy update. Under the uniform prior used in our experiments, $1/p_B(b) = |\mathcal{B}|$ for every budget, so the effective price appearing in the shaped return is $\Lambda_b = |\mathcal{B}|\, \lambda_b$.

\subsection{Gradients of the Lagrangian}
\label{app:gradients}

This section derives the policy and multiplier updates of \Cref{app:optimisation} from the Lagrangian \Cref{eq:lagrangian}, whose per-trajectory integrand is the shaped return $\tilde{R}$ of \baponame{} (\Cref{fig:reward-functions}) and whose trailing term $\delta \sum_{b \in \mathcal{B}} \lambda_b$ is constant in $\pi$.

\paragraph{Policy gradient.}
Fix $\{\lambda_b\}$ and parameterise the policy by $\pi_\theta$ as in \Cref{eq:policy}. In the budget-augmented POMDP $\tilde{\mathcal{M}}$ of \Cref{sec:bapomdp}, neither the kernels $\tilde{\mathcal{T}}, \tilde{\mathcal{O}}$ nor the budget recursion $b_{t+1} = b_t - \kappa(a_t)$ depends on $\theta$, so their log-derivatives vanish and the trajectory score reduces to a sum over the policy's own factors,
\begin{equation}
s_\theta(\zeta, x, B_0) := \nabla_\theta \log p_\theta(\zeta \mid x, B_0) = \sum_{t \leq \tau} \nabla_\theta \log \pi_\theta\!\left(a_t \mid x, h_t, b_t, B_0\right).
\label{eq:score}
\end{equation}
Since $\delta \sum_b \lambda_b$ is constant in $\pi$, the score-function gradient of the Lagrangian is
\begin{equation}
\nabla_\theta \mathcal{L}\big(\pi_\theta, \{\lambda_b\}\big) = \mathbb{E}_{x \sim p_{\mathcal{X}},\, B_0 \sim p_B,\, \zeta \sim \pi_\theta(\cdot \mid x, B_0)}\!\left[\, \tilde{R}(\zeta, B_0)\, s_\theta(\zeta, x, B_0) \right].
\label{eq:policy_gradient}
\end{equation}
Each action $a_t$ is itself a token sequence, so each factor of \Cref{eq:score} expands into per-token log-probabilities and $s_\theta$ sums over all generated tokens. The return is terminal and undiscounted, so \Cref{eq:policy_gradient} weights every token of a trajectory by the same scalar $\tilde{R}$, with no length normaliser.

We reduce variance by centring returns within each group. Write
\[
\tilde{R}^{(g,i)} := \tilde{R}\big(\zeta^{(g,i)}, B_0^{(g)}\big), \qquad
s_\theta^{(g,i)} := s_\theta\big(\zeta^{(g,i)}, x^{(g)}, B_0^{(g)}\big), \qquad
\bar{\tilde{R}}^{(g)} := G^{-1} \sum_{j=1}^{G} \tilde{R}^{(g,j)}.
\]
Within a group, trajectories are i.i.d.\ \emph{given} the task and budget. Given that pair, $\mathbb{E}_{\zeta \sim \pi_\theta(\cdot \mid x, B_0)}[s_\theta(\zeta, x, B_0)] = 0$, so the cross terms vanish. Each trajectory also appears in its own group mean, leaving a factor of $1 - 1/G$ after averaging over the task and budget draws,
\begin{equation}
\mathbb{E}_{\substack{(x^{(g)}, B_0^{(g)}) \,\sim\, p_{\mathcal{X}} \otimes p_B \\ \zeta^{(g,1)}, \ldots, \zeta^{(g,G)} \,\sim\, \pi_\theta(\cdot \mid x^{(g)}, B_0^{(g)})}}\!\left[ \left(\tilde{R}^{(g,i)} - \bar{\tilde{R}}^{(g)}\right) s_\theta^{(g,i)} \right] = \left(1 - \frac{1}{G}\right) \nabla_\theta \mathcal{L}\big(\pi_\theta, \{\lambda_b\}\big),
\label{eq:baseline_scaling}
\end{equation}
with the constant absorbed into the learning rate. We omit GRPO's standard-deviation and length normalisers of \Cref{eq:grpo-adv}, which do not preserve this gradient (\Cref{app:grpo}).

In practice we replace \Cref{eq:policy_gradient} by the clipped Dr.\ GRPO surrogate \Cref{eq:surrogate}, recovering the advantage \Cref{eq:advantage}. Before the segment filtering of \Cref{app:train-grpo}, its expected gradient agrees with the policy gradient up to the factor $1-1/G$ at $\theta = \theta_{\mathrm{old}}$, where every ratio $r_{t,l}^{(g,i)}(\theta)$ equals one, while clipping controls the off-policy change over repeated optimisation steps on one batch of rollouts (\Cref{app:grpo}).

\paragraph{Multiplier gradient.}
Now fix $\pi$ and let $p_b := \mathbb{P}_\pi\!\left(\bar{C}(\zeta,B_0) > 1 \,\middle|\, B_0 = b\right) = g_b(\pi) + \delta$ be the overspend rate at budget $b$. Differentiating \Cref{eq:lagrangian} in $\lambda_b$, with the penalty written per budget as in \Cref{eq:rearrangement}, gives
\begin{equation}
\frac{\partial \mathcal{L}}{\partial \lambda_b} = -g_b(\pi) = -(p_b - \delta).
\label{eq:dual_gradient}
\end{equation}
Since the multipliers minimise $\mathcal{L}$ subject to $\lambda_b \geq 0$, projected gradient descent in $\lambda_b$ gives the projected dual update of \Cref{alg:bapo},
\begin{equation}
\lambda_b \leftarrow \left[\, \lambda_b - \beta \frac{\partial \mathcal{L}}{\partial \lambda_b} \,\right]_+ = \big[\, \lambda_b + \beta (p_b - \delta) \,\big]_+ .
\label{eq:dual_exact}
\end{equation}
Replacing $p_b$ by the batch estimate $\hat{p}_b$ gives \Cref{eq:dual}. Each multiplier therefore increases when $p_b > \delta$ and decreases when $p_b < \delta$, subject to projection at zero. The batch estimate $\hat{p}_b$ is unbiased for $p_b$ under the current policy. We use it without smoothing across updates to keep the estimate on-policy. By \Cref{eq:dual_gradient} the gradient at budget $b$ does not involve $\lambda_{b'}$ for $b' \neq b$, so all multipliers update in parallel from one batch, as in \Cref{alg:bapo}.

At a fixed point of the exact-rate update away from the upper cap in \Cref{eq:dual}, $\lambda_b > 0$ implies $p_b = \delta$, while $\lambda_b = 0$ implies $p_b \leq \delta$. Thus, a budget with a positive, uncapped multiplier has overspend rate $\delta$.

\subsection{Complementary Roles of the Two Penalties}
\label{app:roles}

Both penalties in the \baponame{} return \Cref{eq:lagrangian} activate on the same event $\{\bar{C}(\zeta, B_0) > 1\}$ but act on different quantities. As a function of the normalised cost $c = \bar{C}(\zeta, b)$ at a fixed task outcome, the rate term $\Lambda_b \mathbf{1}[c > 1]$ is constant on $\{c > 1\}$, whereas $P_{\mathrm{DAPO}}$ of \Cref{eq:softoverlong} has slope $\alpha$ on $(1, 2)$. The rate term alone is therefore indifferent between a marginal and a gross overspend. Conversely, $\alpha = \tfrac{1}{2}$ is fixed and shared across budgets, so the overspend rate it induces at any budget is not controlled, whereas an interior fixed point of the dual update gives $p_b = \delta$ (\Cref{app:gradients}).

What reaches the policy is the group-centred advantage \Cref{eq:advantage}. Fix a group $g$ at budget $b$ and drop the index $g$, writing $R^{(i)}$ for the task reward of rollout $i$, $u^{(i)} := \mathbf{1}[\bar{C}(\zeta^{(i)}, b) > 1]$ for its overspend indicator, $P^{(i)} := P_{\mathrm{DAPO}}(\zeta^{(i)}, b)$ for its magnitude penalty and $\hat{p} := G^{-1} \sum_j u^{(j)}$ for the overspend fraction of the group. Centring each term of the return separately gives
\begin{equation}
A^{(i)} = \underbrace{\big(R^{(i)} - \bar{R}\big)}_{\text{task outcome}} \; - \; \underbrace{\big(P^{(i)} - \bar{P}\big)}_{\text{magnitude}} \; - \; \underbrace{\Lambda_b\big(u^{(i)} - \hat{p}\big)}_{\text{rate}} ,
\label{eq:advantage-decomp}
\end{equation}
with bars denoting group means. Which terms survive depends on how the group splits on the budget. If no rollout overspends, both penalty terms vanish and the group reduces to ordinary task learning whatever $\Lambda_b$ has reached. If every rollout overspends, the rate term cancels however large $\Lambda_b$ is, and only differences in the magnitude penalty distinguish rollouts with the same task outcome. This signal also vanishes if every magnitude penalty has saturated. Without it, a group that also shares its task outcome has zero advantage. Such groups can occur at tight budgets early in training, when every rollout overspends and none succeeds; rate-only shaping then supplies no policy-gradient signal. If compliance is mixed, the rate term separates compliant from overspending rollouts by exactly $\Lambda_b$ however far the overspenders ran over, while the magnitude term orders the overspending rollouts by at most $\alpha$. The dual therefore sets the price of overspending at all, adjusted per budget towards the target rate $\delta$, and $P_{\mathrm{DAPO}}$ orders overspending trajectories by how far they ran over. Neither is recoverable from the other by rescaling, so \baponame{} keeps both.

\subsection{Group-Relative Policy Optimisation}
\label{app:grpo}

GRPO \citep{shao2024deepseekmath} is a critic-free variant of PPO \citep{schulman2017proximal}. For each prompt, it samples a group of $G \in \mathbb{Z}_{>0}$ responses with rewards $\{r^{(g,i)}\}_{i=1}^{G} \subset \mathbb{R}$ and uses the group mean as a baseline. For each group $g$, its advantage is
\begin{equation}
A^{(g,i)}
= \frac{1}{\sigma^{(g)}}\left(
r^{(g,i)} - \frac{1}{G}\sum_{j=1}^{G}r^{(g,j)}
\right),
\label{eq:grpo-adv}
\end{equation}
where $\sigma^{(g)}$ is the standard deviation of the rewards within group $g$. This expression applies to groups with nonzero reward variance. GRPO applies this advantage to each response token through a clipped PPO surrogate, averaging the token-level objective within each response. For a batch of $K$ groups with $G$ responses per group, this objective is
\begin{equation}
\mathcal{J}_{\mathrm{GRPO}}(\theta)
= \frac{1}{KG}\sum_{g=1}^{K}\sum_{i=1}^{G}
\frac{1}{N^{(g,i)}}
\sum_{t\leq\tau^{(g,i)}}\sum_{l\leq L_t^{(g,i)}}
\min\!\left(
r_{t,l}^{(g,i)}(\theta)A^{(g,i)},
\operatorname{clip}\!\left(r_{t,l}^{(g,i)}(\theta),1-\epsilon,1+\epsilon\right)A^{(g,i)}
\right),
\label{eq:grpo-surrogate}
\end{equation}
where $N^{(g,i)} := \sum_{t\leq\tau^{(g,i)}} L_t^{(g,i)}$ is the total number of generated tokens in trajectory $(g,i)$, and $r_{t,l}^{(g,i)}(\theta)$ is the token-level importance ratio between the current and behaviour policies.

\citet{liu2025understanding} identify biases from both normalisers in these equations. Dividing by $\sigma$ gives greater weight to groups with lower reward variance. Dividing the surrogate by response length gives shorter positive-advantage responses larger per-token updates, while giving longer negative-advantage responses smaller per-token updates. This can encourage failed responses to grow longer. Dr.\ GRPO removes both normalisers, using
\begin{equation}
A^{(g,i)} = r^{(g,i)} - \frac{1}{G}\sum_{j=1}^{G}r^{(g,j)}
\label{eq:drgrpo-adv}
\end{equation}
and summing the clipped surrogate over response tokens without dividing by response length. The batch averaging factor $1/(KG)$ remains.

We use Dr.\ GRPO throughout, grouping trajectories by task and budget so that the within-group baseline compares rollouts subject to the same budget. \Cref{app:updates} gives the resulting policy update for our shaped return.

\subsection{Policy and Multiplier Updates}
\label{app:optimisation}
\label{app:updates}

\paragraph{Policy update.}
At each iteration, we draw $K \in \mathbb{Z}_{>0}$ task-budget pairs $(x^{(g)}, B_0^{(g)}) \sim p_{\mathcal{X}} \otimes p_B$, $g=1,\ldots,K$, and sample $G \in \mathbb{Z}_{>0}$ trajectories $\zeta^{(g,i)} \sim \pi_\theta(\cdot \mid x^{(g)}, B_0^{(g)})$, $i=1,\ldots,G$, for each pair. Writing $\tilde{R}^{(g,i)} := \tilde{R}(\zeta^{(g,i)}, B_0^{(g)}) \in \mathbb{R}$ for the chosen training return, we centre it within each group:
\begin{equation}
A^{(g,i)} = \tilde{R}^{(g,i)} - \frac{1}{G}\sum_{j=1}^{G}\tilde{R}^{(g,j)}.
\label{eq:advantage}
\end{equation}
Each group therefore contains rollouts for one task at one budget. For \ratename{} and \baponame{}, the group also shares the same multiplier $\lambda_{B_0^{(g)}}$. Centring within groups compares trajectories that faced identical budgets. This is \Cref{eq:drgrpo-adv} with $r^{(i)}$ replaced by $\tilde{R}^{(g,i)}$.

At step $t$, collect the policy's inputs into $z_t^{(g,i)} := \big(x^{(g)}, h_t^{(g,i)}, b_t^{(g,i)}, B_0^{(g)}\big)$ and write the action as its token sequence $a_t^{(g,i)} = \big(a_{t,1}^{(g,i)}, \dots, a_{t,L_t^{(g,i)}}^{(g,i)}\big)$. The token-level importance ratio is
\[
r_{t,l}^{(g,i)}(\theta)
=
\frac{
\pi_\theta\!\big(a_{t,l}^{(g,i)} \mid z_t^{(g,i)}, a_{t,<l}^{(g,i)}\big)
}{
\pi_{\theta_{\mathrm{old}}}\!\big(a_{t,l}^{(g,i)} \mid z_t^{(g,i)}, a_{t,<l}^{(g,i)}\big)
}.
\]
For clip range $\epsilon \in (0,1)$, we ascend the Dr.\ GRPO surrogate
\begin{equation}
\mathcal{J}(\theta)
= \frac{1}{KG}\sum_{g=1}^{K}\sum_{i=1}^{G}
\sum_{t\leq\tau^{(g,i)}}\sum_{l\leq L_t^{(g,i)}}
\min\!\left(
r_{t,l}^{(g,i)}(\theta)A^{(g,i)},
\operatorname{clip}\!\left(r_{t,l}^{(g,i)}(\theta),1-\epsilon,1+\epsilon\right)A^{(g,i)}
\right).
\label{eq:surrogate}
\end{equation}
As in Dr.\ GRPO, the trajectory-level advantage is applied to every token without a response-length normaliser. This policy update applies to every reward objective. For \baponame{}, \Cref{app:gradients} relates it to the policy gradient of \Cref{eq:lagrangian}.

\paragraph{Multiplier update.}
For \ratename{} and \baponame{}, each multiplier $\lambda_b$ increases when the policy overspends at budget $b$ more often than the permitted rate $\delta$, and decreases otherwise. We estimate the overspend rate from the batch rollouts collected at budget $b$, pooled over their tasks. Let
\[
\bar{C}^{(g,i)} := \bar{C}\big(\zeta^{(g,i)},B_0^{(g)}\big),
\qquad
\mathcal{I}_b := \{(g,i) : B_0^{(g)} = b\},
\]
and let $\beta \in \mathbb{R}_{>0}$ be the multiplier step size. The batch estimate and projected update are
\begin{equation}
\hat{p}_b
= \frac{1}{|\mathcal{I}_b|}
\sum_{(g,i)\in\mathcal{I}_b}
\mathbf{1}\!\left[\bar{C}^{(g,i)} > 1\right],
\qquad
\lambda_b \leftarrow
\min\!\left\{
\big[\lambda_b+\beta(\hat{p}_b-\delta)\big]_+,\,
\Lambda_{\max}p_B(b)
\right\}.
\label{eq:dual}
\end{equation}
For budgets with $|\mathcal{I}_b|>0$, this is projected gradient descent on the Lagrangian in $\lambda_b$, whose gradient at the true overspend rate $p_b$ is $\delta-p_b$ (\Cref{app:gradients}). Budgets absent from the batch retain their previous multiplier. The raw batch estimate is not smoothed across updates, keeping the multiplier update on-policy.

The effective-price cap $\Lambda_{\max}$ depends on the reward. For \ratename{}, which has no magnitude penalty, we set $\Lambda_{\max}=1$. For \baponame{}, we set $\Lambda_{\max}=\tfrac{1}{2}$ because the DAPO penalty is also bounded by $\tfrac{1}{2}$. In each case, the cap ensures that an overspending success is not valued below a compliant failure. We project the multiplier itself, rather than clipping only the price in the shaped return, so that the price can decrease when the overspend rate falls below $\delta$. The cap makes this a bounded-price relaxation: a budget can retain an overspend rate above $\delta$ when its price reaches the cap, and we do not claim constraint satisfaction or convergence for the parameterised policy.

\subsection{Training Algorithms}
\label{app:train-algo}

\Cref{alg:budget-training} optimises \Cref{eq:budget-objective} for a chosen fixed return $\tilde{R}$. For simplicity, both algorithms sample budgets uniformly from $\mathcal{B}$, as in our experiments. Rollouts receive budget feedback and end on agent or environment termination, or at $T_{\max}$. \grponormalname{} sets $\tilde{R}(\zeta,B_0)=R(\zeta)$, while the fixed budget-shaped objectives use the returns in \Cref{app:rewards}. \truncname{} also applies its budget cutoff. Both algorithms use the segment filtering described in \Cref{app:train-grpo}.

\begin{algorithm}[ht!]
\caption{Budget-aware training with a fixed reward objective}
\label{alg:budget-training}
\small
\begin{algorithmic}[1]
\Require Policy $\pi_\theta$; task distribution $p_{\mathcal{X}}$; budget set $\mathcal{B}$; cost map $\kappa$; fixed return $\tilde{R}$; groups $K$; group size $G$; clip range $\epsilon$; horizon $T_{\max}$
\For{each training iteration}
    \State $\theta_{\mathrm{old}} \gets \theta$
    \State Draw $K$ pairs $\big(x^{(g)}, B_0^{(g)}\big) \sim p_{\mathcal{X}} \otimes \operatorname{Unif}(\mathcal{B})$
    \For{each group $g$}
        \For{each rollout $i = 1, \ldots, G$}
            \State Reset the environment for $x^{(g)}$ and set $b_1^{(g,i)} \gets B_0^{(g)}$
            \State Collect $\zeta^{(g,i)}$ under $\pi_{\theta_{\mathrm{old}}}$ with budget feedback and the update in \Cref{eq:budget-update}
            \State Compute $R^{(g,i)} \in \{0,1\}$ and $\bar{C}^{(g,i)} \in \mathbb{R}_{\geq0}$ from $\zeta^{(g,i)}$ (\Cref{eq:normcost})
            \State $\tilde{R}^{(g,i)} \gets \tilde{R}(\zeta^{(g,i)}, B_0^{(g)})$
        \EndFor
        \State $A^{(g,i)} \gets \tilde{R}^{(g,i)} - \frac{1}{G}\sum_{j=1}^{G}\tilde{R}^{(g,j)}$ for each $i$
    \EndFor
    \State Apply segment filtering (\Cref{app:train-grpo})
    \State Update $\theta$ using the clipped Dr.\ GRPO surrogate in \Cref{eq:surrogate}
\EndFor
\end{algorithmic}
\end{algorithm}

\Cref{alg:bapo} covers \ratename{} and \baponame{} by adding a learned violation price and updating it from the same rollouts. Under uniform budget sampling, the effective price is $\Lambda_b=|\mathcal{B}|\lambda_b$.

\begin{algorithm}[ht!]
\caption{Budget-aware training with learned violation prices (\ratename{} and \baponame{})}
\label{alg:bapo}
\small
\begin{algorithmic}[1]
\Require Policy $\pi_\theta$; task distribution $p_{\mathcal{X}}$; budget set $\mathcal{B}$; cost map $\kappa$; objective \ratename{} or \baponame{}; tolerance $\delta$; dual step $\beta$; groups $K$; group size $G$; clip range $\epsilon$; horizon $T_{\max}$
\State Set $(\alpha,\Lambda_{\max}) \gets (0,1)$ for \ratename{} or $(\tfrac12,\tfrac12)$ for \baponame{}
\State Initialise $\lambda_b \gets 0$ for each $b \in \mathcal{B}$
\For{each training iteration}
    \State $\theta_{\mathrm{old}} \gets \theta$
    \State Draw $K$ pairs $\big(x^{(g)}, B_0^{(g)}\big) \sim p_{\mathcal{X}} \otimes \operatorname{Unif}(\mathcal{B})$
    \For{each group $g$}
        \For{each rollout $i = 1, \ldots, G$}
            \State Reset the environment for $x^{(g)}$ and set $b_1^{(g,i)} \gets B_0^{(g)}$
            \State Collect $\zeta^{(g,i)}$ under $\pi_{\theta_{\mathrm{old}}}$ with budget feedback and the update in \Cref{eq:budget-update}
            \State Compute $R^{(g,i)} \in \{0,1\}$ and $\bar{C}^{(g,i)} \in \mathbb{R}_{\geq0}$ from $\zeta^{(g,i)}$ (\Cref{eq:normcost})
            \State $\tilde{R}^{(g,i)} \gets R^{(g,i)} - \alpha\operatorname{clip}(\bar{C}^{(g,i)}-1,0,1) - |\mathcal{B}|\lambda_{B_0^{(g)}}\mathbf{1}[\bar{C}^{(g,i)}>1]$
        \EndFor
        \State $A^{(g,i)} \gets \tilde{R}^{(g,i)} - \frac{1}{G}\sum_{j=1}^{G}\tilde{R}^{(g,j)}$ for each $i$
    \EndFor
    \State Apply segment filtering (\Cref{app:train-grpo})
    \State Update $\theta$ using the clipped Dr.\ GRPO surrogate in \Cref{eq:surrogate}
    \For{each budget $b$ sampled in the batch}
        \State $\mathcal{I}_b \gets \{(g,i):B_0^{(g)}=b\}$
        \State $\hat{p}_b \gets \frac{1}{|\mathcal{I}_b|}\sum_{(g,i)\in\mathcal{I}_b}\mathbf{1}[\bar{C}^{(g,i)}>1]$
        \State $\lambda_b \gets \min\{[\lambda_b+\beta(\hat{p}_b-\delta)]_+,\,\Lambda_{\max}/|\mathcal{B}|\}$
    \EndFor
\EndFor
\end{algorithmic}
\end{algorithm}

\FloatBarrier
\section{Experimental Setup}
\label{app:datasets}

We describe the environments, data splits, cost maps, budget grids, and budget-conditioning instructions used in our experiments. Each environment uses the budget-augmented POMDP of \Cref{sec:bapomdp}. The total budget $B_0$ is given at the start of the episode, and the remaining budget $b_t$ is reported after each turn. All evaluations use four rollouts per evaluation instance and budget, while training uses groups of eight rollouts.

\subsection{AppWorld}
\label{app:data-appworld}

AppWorld \citep{trivedi2024appworld} is an interactive coding-agent benchmark in which the agent completes digital tasks by writing Python code that calls simulated app APIs. Each executed code block is one model turn, including the terminal \texttt{apis.supervisor.complete\_task()} call.

We train on the AppWorld training split and evaluate on the $168$ tasks in \texttt{test-normal}. With four rollouts per task, this gives $672$ rollouts per budget. We also evaluate under distribution shift on the $417$ tasks in \texttt{test-challenge}. AppWorld labels each task as easy, medium, or hard, which we use to split spend by difficulty in \Cref{fig:spend}(a). Unless stated otherwise, uncertainty bands for binary success and compliance rates show $\pm1$ standard deviation of the binomial rate estimate.

Episodes have a horizon of $T_{\max}=40$ turns for Qwen3.5-9B and $T_{\max}=50$ turns for Qwen3.5-4B. These horizons exceed the largest training step budgets, allowing overspending to remain observable rather than being hidden by the episode horizon.

\subsection{ASTRA}
\label{app:data-astra}

ASTRA is a personal-assistant benchmark built on ToolSandbox \citep{lu2025toolsandbox}, with tasks involving email, calendars, messages, and contacts. Its native interface uses structured tool calls, whereas AppWorld uses Python code. We expose all ASTRA tools through a single \texttt{all} namespace and retain AppWorld's Python-REPL interface. Calls therefore take the form \texttt{apis.all.api\_name(kwarg=val)}. This keeps the action interface aligned with AppWorld for the transfer experiments of \Cref{sec:exp-ood}. ASTRA grades success by similarity to a set of reference milestones. We binarise this score at a threshold of $0.8$ for our evaluation.

We use ASTRA only to evaluate the AppWorld-trained policies without retraining. The evaluation covers $3116$ ASTRA scenarios under step budgets from $2$ to $14$, with a horizon of $40$ turns. It uses the AppWorld step-budget block of \Cref{code:budget-block-steps}. The ASTRA task instruction retains the AppWorld REPL framing and general instructions where possible and differs only where required by the ASTRA tool interface. \Cref{code:appworld-prompt,code:astra-prompt} give both task instructions in full.

\subsection{BabyAI}
\label{app:data-babyai}

We use a text interface to BabyAI \citep{chevalier-boisvert2019babyai}, a partially observed gridworld environment built on MiniGrid. The agent receives textual observations and selects moves from a small fixed vocabulary, including \emph{turn left}, \emph{turn right}, \emph{move forward}, \emph{pick up}, \emph{drop}, and \emph{toggle}. This provides a task and environment type distinct from the multi-turn coding settings of AppWorld and ASTRA.

We budget BabyAI by model turns rather than individual moves. Each \texttt{action} block may contain multiple moves and costs one unit regardless of how many moves it contains. This parallels the AppWorld step budget, where an agent can batch multiple API calls within a single turn. The moves execute sequentially without intermediate observations. An invalid move invalidates the entire block while still consuming the turn. The agent can therefore conserve budget by batching moves, at the cost of acting further ahead without observing intermediate states.

We use $15$ BabyAI levels covering a range of navigation and object-interaction tasks. Each level defines a family of related tasks rather than a single fixed environment. We generate concrete task instances from each level using fixed random seeds. A seed determines the mission, environment layout, object and door placement, and the agent's initial state.

We construct non-overlapping training and evaluation sets using disjoint seed pools. No concrete task instance therefore appears in both splits, although both cover the same level families. The evaluation seeds are fixed and shared across methods, so every policy is evaluated on the same missions and environment configurations.

\subsection{Cost Maps and Budgeted Quantities}
\label{app:data-costmaps}

We vary the budgeted quantity through the cost map $\kappa$ while leaving the rest of the training framework unchanged. \Cref{tab:costmaps} gives the three cost maps used in our experiments.

\begin{table}[htbp]
\centering
\small
\setlength{\tabcolsep}{5pt}
\renewcommand{\arraystretch}{1.15}
\caption{\textbf{Budgeted quantities.} Quantities defined by the cost map $\kappa$. The domain abbreviations denote AppWorld, ASTRA, and BabyAI.}
\label{tab:costmaps}
\begin{tabularx}{\linewidth}{@{}l l >{\raggedright\arraybackslash}X c@{}}
\toprule
Unit & $\kappa(a_t)$ & Interpretation of $B_0$ & Domains \\
\midrule
Steps & $1$ & Model turns available to the agent & \textsc{aw}, \textsc{as}, \textsc{ba} \\
Tool calls & \# API calls in $a_t$ & API calls available to the agent & \textsc{aw} \\
Tokens & $|a_t|$ in tokens & Generated tokens, including reasoning and actions & \textsc{aw} \\
\bottomrule
\end{tabularx}
\end{table}

Under a step budget, each model turn costs one unit. On BabyAI, one model turn may contain several primitive moves. Under a tool-call budget, each API call is charged independently. Under a token budget, all tokens generated by the policy are charged, including its reasoning and actions. Tokens in the task instruction and environment observations are not charged.

\paragraph{Budget grids.}
On AppWorld, we construct a separate six point training grid for each model and cost unit. The grid is based on the empirical resource use of \grponormalname{}, trained with the shared Dr.\ GRPO pipeline and a budget-independent task reward. We select budgets at the $1$st, $10$th, $25$th, $50$th, $75$th, and $90$th percentiles of its spend distribution. This produces budgets spanning the observed range of task demands, from highly constrained settings to comparatively generous ones. All reward formulations use the same training and evaluation grids within each model and cost unit.

All AppWorld evaluation grids include training budgets and unseen budgets between the training points. The step and API-call grids also include budgets below and above the training range; the token grids remain within it. \Cref{tab:appworld-grids} gives the grids. For BabyAI, all methods and both model scales use the same six turn budgets, $\{5, 7, 11, 16, 23, 34\}$, for training and evaluation.

\begin{table}[H]
\centering
\scriptsize
\setlength{\tabcolsep}{4pt}
\renewcommand{\arraystretch}{1.25}
\caption{\textbf{Budget grids.} Training and evaluation budget grids by cost unit and model. All methods use the same grids. }

\label{tab:appworld-grids}
\begin{tabularx}{\linewidth}{@{}l l >{\raggedright\arraybackslash}X >{\raggedright\arraybackslash}X@{}}
\toprule
Unit & Model & Training grid $\mathcal{B}$ & Evaluation grid $\mathcal{B}_{\mathrm{eval}}$ \\
\midrule
Steps
 & Qwen3.5-9B     & \{10, 15, 18, 22, 25, 29\} & \{4, 8, 10, 12, 15, 18, 20, 22, 25, 27, 29, 32, 36\} \\
 & Qwen3.5-4B     & \{10, 15, 17, 20, 24, 26\} & \{4, 8, 9, 10, 11, 12, 15, 17, 18, 20, 21, 22, 24, 26, 29, 31, 32, 34, 36\} \\
\addlinespace
Tool calls
 & Qwen3.5-9B     & \{10, 15, 18, 22, 26, 29\} & \{6, 10, 12, 15, 18, 20, 22, 24, 26, 29, 32, 36\} \\
 & Qwen3.5-4B     & \{13, 15, 17, 19, 22, 25\} & \{6, 10, 11, 12, 13, 14, 15, 17, 19, 20, 21, 22, 24, 25, 26, 27, 30, 31, 35, 36\} \\
\addlinespace
Tokens
 & Qwen3.5-9B     & \{896, 1408, 1856, 2432, 3136, 3904\} & \{896, 1152, 1408, 1600, 1856, 2112, 2432, 2752, 3136, 3520, 3904\} \\
 & Qwen3.5-4B     & \{1452, 2073, 2489, 3066, 4264, 5186\} & \{1452, 1728, 2073, 2240, 2489, 2752, 3066, 3584, 4264, 4672, 5186\} \\
\addlinespace
Turns
 & Qwen3.5-4B, 9B & \{5, 7, 11, 16, 23, 34\} & \{5, 7, 11, 16, 23, 34\} \\
\bottomrule
\end{tabularx}
\end{table}

\subsection{Budget Conditioning and Prompt Templates}
\label{app:data-budget}
\label{app:data-prompts}

At the start of each episode, the policy receives a domain-specific task instruction followed immediately by a \texttt{<budget>} block. The block states the total budget $B_0$. After every model turn, the environment appends the remaining budget $b_t$ to the next observation. The policy therefore observes both the original and remaining budgets throughout the episode, as required by \Cref{eq:policy}.

\Cref{code:appworld-prompt} gives the AppWorld task instruction, and \Cref{code:astra-prompt} gives its ASTRA adaptation. AppWorld uses separate budget blocks for step, API-call, and token budgets, which are given in \Cref{code:budget-block-steps,code:budget-block-calls,code:budget-block-tokens}. \Cref{code:babyai-prompt,code:babyai-block} give the BabyAI task instruction and turn-budget block. All models, methods, and splits within a domain and cost unit receive the same text. Only the task-specific fields and total budget vary, together with the user and date fields where applicable.

\begin{pbox}[label={code:appworld-prompt}]{AppWorld Task Instruction}
\footnotesize
I am your supervisor and you are a super intelligent AI Assistant whose job is to achieve my day-to-day tasks completely autonomously.\\[3pt]
To do this, you will need to interact with app/s (e.g., spotify, venmo etc) using their associated APIs on my behalf. For this you will undertake a *multi-step conversation* using a python REPL environment. That is, you will write the python code and the environment will execute it and show you the result, based on which, you will write python code for the next step and so on, until you've achieved the goal. This environment will let you interact with app/s using their associated APIs on my behalf.\\[3pt]
Here are three key APIs that you need to know to get more information\\[3pt]
\# To get a list of apps that are available to you.\\
\texttt{print(apis.api\_docs.show\_app\_descriptions())}\\[2pt]
\# To get the list of apis under any app listed above, e.g. spotify\\
\texttt{print(apis.api\_docs.show\_api\_descriptions(app\_name='spotify'))}\\[2pt]
\# To get the specification of a particular api, e.g. spotify app's login api\\
\texttt{print(apis.api\_docs.show\_api\_doc(app\_name='spotify', api\_name='login'))}\\[3pt]
Each code execution will produce an output that you can use in subsequent calls. Using these APIs, you can now generate code, that I will execute, to solve the task.\\[3pt]
**Key instructions**:\\
(1) Make sure to end code blocks with \texttt{\textasciigrave\textasciigrave\textasciigrave} followed by a newline(\textbackslash n).\\
(2) Remember you can use the variables in your code in subsequent code blocks.\\
(3) You can use the ``supervisor'' app to get information about my accounts and use the ``phone'' app to get information about friends and family.\\
(4) Always look at API specifications (using \texttt{apis.api\_docs.show\_api\_doc}) before calling an API.\\
(5) Write small chunks of code and only one chunk of code in every step. Make sure everything is working correctly before making any irreversible change.\\
(6) Many APIs return items in ``pages''. Make sure to run through all the pages by looping over \texttt{page\_index}.\\
(7) Once you have completed the task, make sure to call \texttt{apis.supervisor.complete\_task()}. If the task asked for some information, return it as the answer argument, i.e. call \texttt{apis.supervisor.complete\_task(answer=<answer>)}. Many tasks do not require an answer, so in those cases, just call \texttt{apis.supervisor.complete\_task()} i.e. do not pass any argument.\\
(8) Do exactly what the task asks for, nothing more. Do not create, modify, or delete any data that the task does not explicitly require. If the task only asks you to read information, do not write or modify anything. If the task asks you to perform a specific action, perform only that action using the most direct approach available.\\
(9) If the task mentions a message, email, or request from someone, always read the actual content of that message/email first --- it may contain important details, constraints, or filters not stated in the task description.\\
(10) When a task depends on the current date, time, or day of the week, always determine it programmatically (e.g., \texttt{from datetime import datetime; datetime.now()}) --- never guess or assume.\\
(11) When filtering entities by category or relationship (e.g., ``friends'', ``coworkers''), verify membership using the primary/authoritative data source for that relationship --- do not infer it from other app-specific lists.\\
(12) When a task references a time range like ``last N days including today'', today counts as day 1 --- so the start date is N-1 days ago, not N days ago.\\
(13) When writing structured output files (CSV, etc.), use minimal formatting --- do not add unnecessary quoting or encoding around values.\\
(14) You should follow three phases: First, explore the APIs and all relevant data. You can explore a bit more alternative data just to go sure that you did not miss anything. Second, summarize your findings and compare them to the task again. Think about whether you might have missed something. Look it up and verify. Third, once you have all information, start conducting the task and complete it. Before you finish, verify that every aspect of the task has been accounted for.\\[3pt]
Using these APIs, now generate code to solve the actual task:\\[3pt]
My name is: \tvar{main\_user.first\_name} \tvar{main\_user.last\_name}. My personal email is \tvar{main\_user.email} and phone number is \tvar{main\_user.phone\_number}. The current date is \tvar{main\_user.session\_start\_time}.\\
Task: \tvar{input\_str}\\[3pt]
\pslot{budget-conditioning block, appended here}
\end{pbox}

\begin{pbox}[label={code:astra-prompt}]{ASTRA Task Instruction, aligned with \Cref{code:appworld-prompt}}
\footnotesize
I am your supervisor and you are a super intelligent AI Assistant whose job is to achieve my day-to-day tasks completely autonomously.\\[3pt]
To do this, you will need to interact with apps using their associated APIs on my behalf. For this you will undertake a *multi-step conversation* using a python REPL environment. That is, you will write the python code and the environment will execute it and show you the result, based on which, you will write python code for the next step and so on, until you've achieved the goal. This environment will let you interact with these APIs on my behalf.\\[3pt]
Here are three key APIs that you need to know to get more information\\[3pt]
\# To get a list of apps that are available to you (this only ever lists the single 'all' app).\\
\texttt{print(apis.api\_docs.show\_app\_descriptions())}\\[2pt]
\# To get the list of apis under the 'all' app.\\
\texttt{print(apis.api\_docs.show\_api\_descriptions(app\_name='all'))}\\[2pt]
\# To get the specification of a particular api, e.g. the 'all' app's send\_email api\\
\texttt{print(apis.api\_docs.show\_api\_doc(app\_name='all', api\_name='send\_email'))}\\[3pt]
Each code execution will produce an output that you can use in subsequent calls. Using these APIs, you can now generate code, that I will execute, to solve the task.\\[3pt]
**Key instructions**:\\
(1) Make sure to end code blocks with \texttt{\textasciigrave\textasciigrave\textasciigrave} followed by a newline(\textbackslash n).\\
(2) Remember you can use the variables in your code in subsequent code blocks.\\
(3) There is exactly one app, named 'all', which holds every task API. Use the \texttt{apis.api\_docs.*} calls only to look up documentation, and call every task API as \texttt{apis.all.api\_name(kwarg1=val1, ...)}. Do not invent other app names, and do not paste documentation output back as code.\\
(4) Always look at API specifications (using \texttt{apis.api\_docs.show\_api\_doc}) before calling an API.\\
(5) Call APIs using only keyword arguments like \texttt{apis.all.api\_name(kwarg1=val1, ...)}\\
(6) Write small chunks of code and only one chunk of code in every step. Make sure everything is working correctly before making any irreversible change.\\
(7) Many APIs return items in ``pages''. Make sure to run through all the pages by looping over \texttt{page\_index}.\\
(8) Once you have completed the task, make sure to call \texttt{apis.supervisor.complete\_task()}. If the task asked for some information, return it as the answer argument, i.e. call \texttt{apis.supervisor.complete\_task(answer=<answer>)}. Many tasks do not require an answer, so in those cases, just call \texttt{apis.supervisor.complete\_task()} i.e. do not pass any argument.\\
(9) Do exactly what the task asks for, nothing more. Do not create, modify, or delete any data that the task does not explicitly require. If the task only asks you to read information, do not write or modify anything. If the task asks you to perform a specific action, perform only that action using the most direct approach available.\\
(10) If the task mentions a message, email, or request from someone, always read the actual content of that message/email first --- it may contain important details, constraints, or filters not stated in the task description.\\
(11) When a task depends on the current date, time, or day of the week, always determine it programmatically --- never guess or assume.\\
(12) When filtering entities by category or relationship (e.g., ``friends'', ``coworkers''), verify membership using the primary/authoritative data source for that relationship --- do not infer it from other app-specific lists.\\
(13) When a task references a time range like ``last N days including today'', today counts as day 1 --- so the start date is N-1 days ago, not N days ago.\\
(14) When writing structured output files (CSV, etc.), use minimal formatting --- do not add unnecessary quoting or encoding around values.\\
(15) You should follow three phases: First, explore the APIs and all relevant data. You can explore a bit more alternative data just to go sure that you did not miss anything. Second, summarize your findings and compare them to the task again. Think about whether you might have missed something. Look it up and verify. Third, once you have all information, start conducting the task and complete it. Before you finish, verify that every aspect of the task has been accounted for.\\[3pt]
Using these APIs, now generate code to solve the actual task:\\[3pt]
The current date is \tvar{session\_start\_time}.\\
Task: \tvar{input\_str}\\[3pt]
\pslot{budget-conditioning block, appended here}
\end{pbox}

\begin{pbox}[label={code:budget-block-steps}]{Budget-conditioning block for step budgets on AppWorld and ASTRA}
\footnotesize
\btag{<budget>}You have a budget of $B_0$ steps.\\[2pt]
Only the code blocks you send count toward your step budget. After each step you are told how many steps remain.\\[2pt]
Before sending a code block, think carefully about how many steps you have left and plan how best to use them to complete the task within budget. Because a whole block costs just one step, batch several API calls into a single block whenever they don't depend on each other's results --- making one call per step wastes budget. Only split work across steps when a later call genuinely needs an earlier call's output.\\[2pt]
Solve the task successfully while staying strictly within your allocated step budget.\btag{</budget>}
\end{pbox}

\begin{pbox}[label={code:budget-block-calls}]{Budget-conditioning block for API-call budgets on AppWorld}
\footnotesize
\btag{<budget>}You have a budget of $B_0$ API calls.\\[2pt]
Every \texttt{apis.<app>.<endpoint>()} call you make costs 1 from your budget --- this includes \texttt{apis.api\_docs} lookups, not just calls to the task apps themselves. Thinking and plain code are free, and it doesn't matter how many turns your calls are spread across: batching several into one turn costs the same as spreading them out --- only making fewer calls total reduces cost. After each step you are told how many API calls remain.\\[2pt]
Before making a call, check whether you already have what you need from an earlier call, an earlier doc lookup, or the task description --- don't re-look up something you've already seen this episode, and don't make a call just in case. Plan which calls are actually necessary to solve the task.\\[2pt]
Solve the task successfully while staying strictly within your allocated API-call budget.\btag{</budget>}
\end{pbox}

\begin{pbox}[label={code:budget-block-tokens}]{Budget-conditioning block for token budgets on AppWorld}
\footnotesize
\btag{<budget>}You have a budget of $B_0$ tokens.\\[2pt]
Every token you generate counts toward your token budget: your reasoning between \texttt{<think>} and \texttt{</think>}, plus the code you write after \texttt{</think>}. The task description and the environment's replies are free --- only your own output is charged.\\[2pt]
After each step you are told how many tokens you have left and how many your previous step used, so you can calibrate as you go. Before each step, decide the total number of tokens that step will use across your reasoning and your code, state that number in your first line, and keep to it --- think enough to act well but no more, since verbose reasoning spends budget just like long code. Then re-plan at the next step using the updated count.\\[2pt]
Batch independent API calls into one block whenever they don't depend on each other's results; only split work across steps when a later call needs an earlier call's output.\\[2pt]
Solve the task successfully while staying strictly within your allocated token budget.\btag{</budget>}
\end{pbox}

\begin{pbox}[label={code:babyai-prompt}]{BabyAI task instruction}
\footnotesize
You are an agent in a 2D gridworld. You see only a $7\times7$ area in front of you, and parts of the grid remain hidden until you explore them. At each turn, you receive a text description of what you can currently see and the direction you face. You reply with one or more moves. The available moves are listed below.\\[3pt]
\texttt{left} turns left by $90$ degrees without moving.\\
\texttt{right} turns right by $90$ degrees without moving.\\
\texttt{forward} moves one tile forward.\\
\texttt{pick} picks up the object directly in front of you.\\
\texttt{drop} drops the object you are carrying.\\
\texttt{toggle} opens, closes, or activates the object directly in front of you.\\[3pt]
Give one or more moves in a single action block, with one move per line. The moves execute sequentially from your current state. An unrecognised move invalidates the entire block. None of its moves are executed, but the turn is still consumed.\\[3pt]
Task instruction \tvar{mission}\\[3pt]
\pslot{budget-conditioning block, appended here}
\end{pbox}

\begin{pbox}[label={code:babyai-block}]{Budget-conditioning block for BabyAI turn budgets}
\footnotesize
\btag{<budget>}You have a budget of $B_0$ turns. After each turn, you will be told how many turns remain.\\[2pt]
Each action block you send costs one turn. You may batch several moves into one action block when you are confident about the sequence. The moves execute sequentially without intermediate observations, so use a new turn when you need to observe the result before deciding what to do next.\\[2pt]
Solve the task successfully while staying strictly within your allocated turn budget.\btag{</budget>}
\end{pbox}

\noindent
At every subsequent turn, the environment appends the remaining budget as \texttt{<budget>Budget remaining $b_t$ of $B_0$ \{\{ unit \}\}.</budget>}, with the unit matching the cost map. Once the policy exceeds its budget, the message reports zero remaining and states how far the policy is over budget. Under a token budget, it also reports the tokens used by the preceding turn and repeats the request to estimate the length of the next turn.

The guidance differs by resource type because each budget can be conserved in a different way. Batching API calls into one code block conserves steps but does not conserve API calls. Token budgets charge both reasoning and code. BabyAI allows several primitive moves to be placed in one turn, but the agent receives no intermediate observation between those moves.

\subsection{Models and Dr.\ GRPO Settings}
\label{app:train-grpo}

We train Qwen3.5 at 4B and 9B parameters and report both scales throughout. Every method continues training from each model's released post-trained checkpoint, with no supervised warm start of our own. All parameters are updated, with no LoRA, adapters, or frozen layers.

Optimisation uses the group-centred Dr.\ GRPO advantage \Cref{eq:advantage}, with neither length nor standard-deviation normalisation, at $\gamma = 1$, for $120$ updates on Qwen3.5-9B and $80$ on Qwen3.5-4B. The learning rate is flat at $3\times10^{-6}$ with no schedule. Thinking is enabled, with a $768$-token allowance on Qwen3.5-9B and $512$ on Qwen3.5-4B. Under the step and tool-call cost maps these tokens do not count against the budget, whereas under the token cost map of \Cref{app:data-costmaps} they are priced.

Each iteration draws $K$ groups (\Cref{eq:advantage}), with $K = 8$ on Qwen3.5-9B and $K = 32$ on Qwen3.5-4B, each comprising $G = 8$ completions for a shared task and a shared budget, sampled at temperature $1.0$ with nucleus and top-$k$ truncation disabled. The policy update is the clipped surrogate \Cref{eq:surrogate} with symmetric clip range $\epsilon = 0.2$ and no KL penalty or entropy bonus, so the return is shaped only by each method's own objective (\Cref{sec:shaped}). Qwen3.5-9B takes $2$ epochs per update with gradient accumulation over $8$ microbatches of at most $4$ segments, and Qwen3.5-4B takes $1$ epoch over $2$ microbatches of at most $8$. Before the update, groups whose advantages are all zero are dropped, and negative-advantage segments are subsampled so that they make up at most a quarter of the batch. This additional sampling step changes the relative weighting of positive and negative advantages; the unfiltered gradient calculation in \Cref{app:gradients} does not establish unbiasedness of the implemented update.

\subsection{Budget and Constraint Settings}
\label{app:train-budget}

Budgets are drawn uniformly from $\mathcal{B}$, so $1 / p_B(b) = |\mathcal{B}|$ and the effective price of \baponame{} and \ratename{} (\Cref{fig:reward-functions}) is $\Lambda_b = |\mathcal{B}| \lambda_b$. The AppWorld horizons of \Cref{app:data-appworld} exceed the largest training step budgets, so step-budget overspending remains observable to both the dual and soft-overlong penalties. The ramp of \Cref{eq:softoverlong} saturates at twice the budget, so under the step cost map it reaches its maximum of $\alpha$ only when $2b \leq T_{\max}$. On AppWorld, the top three budgets of the Qwen3.5-9B step grid and the top budget of the Qwen3.5-4B step grid fall outside this range. At these budgets, the penalty grades overspending across the entire reachable range without saturating.

We set the permitted overspend rate to $\delta = 0.05$. The dual step size is $\beta = 0.01$ for Qwen3.5-9B and $\beta = 0.04$ for Qwen3.5-4B. Under the uniform six-rung budget prior, these give effective-price step sizes of $|\mathcal{B}|\beta=0.06$ and $0.24$, respectively. We update the dual using the raw within-batch overspend rate in \Cref{eq:dual}, without smoothing across batches. The soft-overlong coefficient is $\alpha=0.5$, following the reward-scale argument in \Cref{app:dapo}.

Multipliers are initialised at $\lambda_b=0$ for every budget. For \ratename{}, they are projected onto $\big[0,p_B(b)\big]$, giving an effective-price cap of $\Lambda_b\leq1$. For \baponame{}, they are projected onto $\big[0,(1-\alpha)p_B(b)\big]$, giving an effective-price cap of $\Lambda_b\leq1-\alpha=0.5$. These are the largest prices for which an overspending success is not valued below a compliant failure under the corresponding reward formulation.

\begin{table}[ht!]
\centering
\small
\setlength{\tabcolsep}{5pt}
\renewcommand{\arraystretch}{1.15}
\caption{\textbf{Training hyperparameters.} Budget grids are given by model and cost unit in \Cref{tab:appworld-grids}. Settings that differ between model scales or reward formulations are stated explicitly. All other settings are shared across methods and domains.}
\label{tab:hparams}
\begin{tabularx}{\linewidth}{@{}l >{\raggedright\arraybackslash}X@{}}
\toprule
Hyperparameter & Value \\
\midrule
\multicolumn{2}{@{}l}{\emph{Model and optimisation}} \\
\addlinespace[1pt]
Model families and scales & Qwen3.5 4B and 9B (released post-trained checkpoints) \\
Fine-tuning & full-parameter, no LoRA or adapters \\
Optimiser & Dr.\ GRPO, $\gamma = 1$ \\
Training updates & $120$ (Qwen3.5-9B), $80$ (Qwen3.5-4B) \\
Learning rate & flat, no schedule, $3\times10^{-6}$ \\
\addlinespace[3pt]
\multicolumn{2}{@{}l}{\emph{Budget family and constraint}} \\
\addlinespace[1pt]
Budget grid $\mathcal{B}$ & six rungs per model and cost unit (\Cref{tab:appworld-grids}) \\
Budget prior $p_B$ & uniform on $\mathcal{B}$ \\
Episode horizon $T_{\max}$ & AppWorld $40$ (Qwen3.5-9B) and $50$ (Qwen3.5-4B); BabyAI $50$; ASTRA $40$ at evaluation \\
Overspend rate $\delta$ & $0.05$ \\
Dual step $\beta$ & $0.01$ (Qwen3.5-9B), $0.04$ (Qwen3.5-4B) \\
Initial multipliers $\lambda_b$ & $0$ for all $b$, with $\Lambda_b$ capped at $1$ for \ratename{} and at $1-\alpha=0.5$ for \baponame{} \\
Soft-overlong coefficient $\alpha$ & $0.5$ \\
\addlinespace[3pt]
\multicolumn{2}{@{}l}{\emph{Rollout collection and policy update}} \\
\addlinespace[1pt]
Groups per iteration $K$ & $8$ (Qwen3.5-9B), $32$ (Qwen3.5-4B) \\
Group size $G$ & $8$ \\
Sampling temperature & $1.0$ (top-$p$, top-$k$ disabled) \\
Thinking tokens & $768$ (Qwen3.5-9B), $512$ (Qwen3.5-4B) \\
PPO clip range $\epsilon$ & $0.2$ (symmetric) \\
PPO epochs per update & $2$ (Qwen3.5-9B), $1$ (Qwen3.5-4B) \\
Gradient accumulation & $8$ microbatches of $\leq 4$ segments (Qwen3.5-9B), $2$ of $\leq 8$ (Qwen3.5-4B) \\
Gradient-norm clip & $1.0$ \\
Entropy and KL coefficients & $0$, no KL term \\
\bottomrule
\end{tabularx}
\end{table}

\subsection{Compute}
\label{app:train-compute}

Training runs on $8 \times$ B200 GPUs with tensor-parallel vLLM generation and an FSDP trainer. Evaluation runs on $8 \times$ H100 GPUs for Qwen3.5-9B and $4 \times$ H100 for Qwen3.5-4B.

\paragraph{What the dual costs to train.}
\baponame{} and \magname{} provide a direct comparison of the additional algorithmic cost of the dual update. The dual introduces only $|\mathcal{B}|$ scalar multipliers, which are updated once per iteration using \Cref{eq:dual}. It requires no additional model parameters, rollouts, forward passes, or backward passes. Both methods use the same rollout configuration, optimiser settings, and hardware. Their realised compute may nevertheless differ because the learned policies generate trajectories of different lengths and advantage filtering may retain different numbers of segments. We therefore treat the methods as matched in training configuration rather than claiming identical wall-clock or optimiser cost.

\section{Additional Results}
\label{app:results}

\subsection{Across-Budget Utility Metrics}
\label{app:ranking-robustness}

We define the summary metrics reported in \Cref{tab:main-winrate-pooled}. An evaluation setting is a combination of model scale, environment, and budget unit. We consider eight settings. These comprise AppWorld step, tool-call, and token budgets and BabyAI turn budgets for each of Qwen3.5-9B and Qwen3.5-4B.

We compare the ten budget-shaping mechanisms and exclude \basename{} and \grponormalname{} because their rewards do not depend on the budget.

\paragraph{Metrics.}
Let $\mathcal{R}$ denote the set of evaluation settings and let $\mathcal{M}_r$ denote the methods evaluated in setting $r\in\mathcal{R}$. Let $B_r$ be the evaluation budget grid for setting $r$, and let $\mathcal{U}_{m,r}(b)$ denote the observed budget utility of method $m$ at budget $b$. The highest observed utility at each budget is
\begin{equation}
\mathcal{U}^{*}_{r}(b)
=
\max_{m\in\mathcal{M}_r}
\mathcal{U}_{m,r}(b).
\end{equation}

For each method $m$, let $\mathcal{R}_m\subseteq\mathcal{R}$ contain the settings in which it was evaluated. We define
\begin{align}
\mathrm{Win}(m)
&=
\frac{1}{|\mathcal{R}_m|}
\sum_{r\in\mathcal{R}_m}
\frac{1}{|B_r|}
\sum_{b\in B_r}
\mathbf{1}
\left[
\mathcal{U}_{m,r}(b)
=
\mathcal{U}^{*}_{r}(b)
\right],
\label{eq:winrate}
\\
\mathrm{NearBest}_{\epsilon}(m)
&=
\frac{1}{|\mathcal{R}_m|}
\sum_{r\in\mathcal{R}_m}
\frac{1}{|B_r|}
\sum_{b\in B_r}
\mathbf{1}
\left[
\mathcal{U}_{m,r}(b)
\geq
\mathcal{U}^{*}_{r}(b)-\epsilon
\right],
\label{eq:nearbest}
\\
\mathrm{Regret}(m)
&=
\frac{1}{|\mathcal{R}_m|}
\sum_{r\in\mathcal{R}_m}
\frac{1}{|B_r|}
\sum_{b\in B_r}
\left[
\mathcal{U}^{*}_{r}(b)
-
\mathcal{U}_{m,r}(b)
\right].
\label{eq:regret}
\end{align}

Win rate measures how often a method attains the highest observed utility at an individual budget. It is a strict measure and can be sensitive to small differences between methods. Near-best rate measures how often a method remains within $\epsilon$ of the best observed utility and is therefore less sensitive to small estimation differences. Mean regret measures the average utility lost relative to the best observed method and accounts for the magnitude of each difference.

We use $\epsilon=0.02$, corresponding to two percentage points of budget utility. Exact ties count as wins for every tied method, so win rates need not sum to $100\%$. Each metric is first averaged over budgets within a setting and then averaged equally across settings. Settings with denser evaluation grids therefore receive no additional weight. Win and near-best rates are reported as percentages, while regret is reported in percentage points.

\paragraph{Uncertainty.}
Budget evaluations within a setting share a trained policy and overlapping tasks, so individual budget points are not independent replicates. \Cref{tab:main-winrate-pooled} reports descriptive point estimates, without confidence intervals or a test of differences between methods. The separate seed study below assesses training variability for two objectives in one setting; it does not quantify uncertainty for every pooled ranking.

\subsection{Seed Robustness and Longer Training}
\label{app:seed-ablation}

\begin{figure}[t]
\centering
\includegraphics[width=0.8\linewidth]{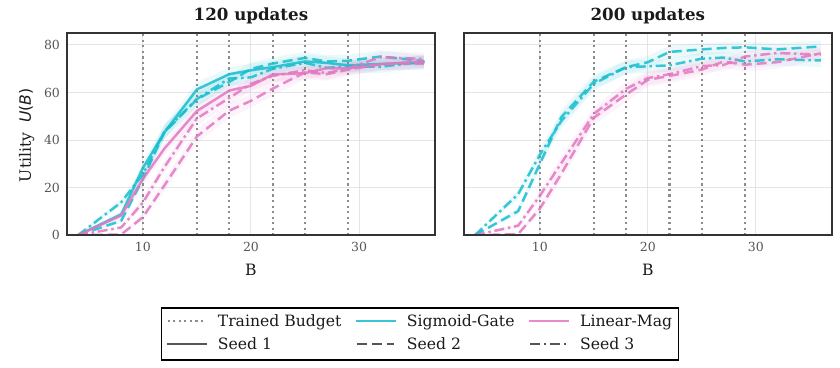}
\caption{\textbf{Seed robustness and longer training.} Budget utility $\mathcal{U}(b)$ on AppWorld \texttt{test-normal} under step budgets for Qwen3.5-9B \sigmoidgatename{} and \magname{}. Two additional seeds are evaluated after $120$ updates (left) and $200$ updates (right), with the main $120$-update seed shown in both panels. Dotted lines mark training budgets. Bands show $\pm1$ standard error over tasks.}
\label{fig:seed-ablation}
\end{figure}

The comparisons in \Cref{sec:exp-objectives} use one training seed per method and setting. To check how much of the spread between methods is training noise, we retrain \sigmoidgatename{} and \magname{} on Qwen3.5-9B with AppWorld step budgets using two further seeds, and continue these runs from the $120$ updates of the main results to $200$ updates. \Cref{fig:seed-ablation} reports their budget utility on \texttt{test-normal}.

After $120$ updates, the three \magname{} seeds differ by around ten percentage points at intermediate budgets, and the \sigmoidgatename{} seeds by a few points. The broad ranking of the two methods is preserved across seeds and with longer training, with \sigmoidgatename{} generally achieving higher utility at tighter and intermediate budgets. The curves become closer and can cross at the loosest budgets, where the budget is less restrictive. Training for $200$ updates improves utility at many budgets.

\subsection{Resource Use on BabyAI}
\label{app:babyai-resource-use}

\begin{figure}[t]
\centering
\includegraphics[width=0.9\linewidth]{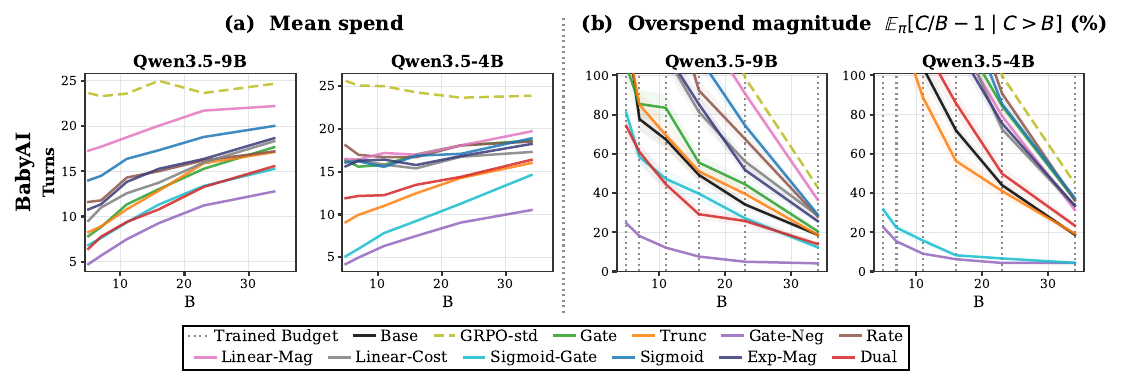}
\caption{\textbf{Resource use on BabyAI.} \textbf{(a)} Mean turns used and \textbf{(b)} mean percentage overspend among rollouts that exceed the budget, for Qwen3.5-9B and 4B. Horizontal axes show the stated turn budget $B$, with dotted lines marking training budgets. Overspend axes are capped at $100\%$.}
\label{fig:babyai-spend-excess}
\end{figure}

Budget utility measures whether a task is completed within budget, but it does not show how a policy adjusts its spending or how severely it exceeds the budget when it fails to comply. \Cref{fig:babyai-spend-excess} therefore reports mean trajectory spend and conditional overspend magnitude across BabyAI turn budgets.

Mean spend is the average number of turns used by the policy. It shows whether resource use changes with the available budget. Lower spend is not necessarily better because using additional turns may improve task success. We define conditional overspend magnitude as
\begin{equation}
\mathrm{Excess}(b)
=
\mathbb{E}
\left[
\bar{C}(\zeta,b)-1
\;\middle|\;
\bar{C}(\zeta,b)>1
\right].
\end{equation}
This is the expected relative excess among trajectories that exceed the budget. Compliance measures how often the policy stays within budget, while $\mathrm{Excess}(b)$ measures the severity of its violations. We report $\mathrm{Excess}(b)$ as a percentage in the figure.

Violations are generally most severe under the tightest budgets and become less severe as the budget grows.

\subsection{Qualitative Examples of Budget Use}
\label{app:qualitative-budget-use}

\begin{tcolorbox}[breakable, enhanced, colback=promptback, colframe=promptframe, title={BabyAI: batching moves under a turn budget}, fonttitle=\sffamily\bfseries\small, coltitle=white, colbacktitle=promptframe, arc=2.5pt, boxrule=0.5pt]
\small
\textbf{Qwen3.5-9B \sigmoidgatename{}, \emph{OpenTwoDoors}, $B=7$.} The policy first plans the route against its stated allowance: ``This should take 5 turns maximum, well within my 7 turn budget.'' It then sends the seven-move trip to the second door in one action block and completes the task in four turns.
\end{tcolorbox}

\begin{tcolorbox}[breakable, enhanced, colback=promptback, colframe=promptframe, title={AppWorld: batching tool operations under a step budget}, fonttitle=\sffamily\bfseries\small, coltitle=white, colbacktitle=promptframe, arc=2.5pt, boxrule=0.5pt]
\small
\textbf{Qwen3.5-9B \sigmoidgatename{} training rollout, $B=15$, two steps remaining.} The policy states: ``I can batch multiple delete calls in one step,'' and combines independent tool operations in a single code block.
\end{tcolorbox}

\FloatBarrier

\end{document}